%% file: Capri_review.tex
\documentclass[10pt,twocolumn,letterpaper]{article}

\usepackage{iccv}
\usepackage{times}
\usepackage{epsfig}
\usepackage{graphicx}
\usepackage{amsmath}
\usepackage{amssymb,bm}
\usepackage{booktabs}
\usepackage{mathrsfs}
\usepackage[T1]{fontenc}
\usepackage{textcomp}
\usepackage{bbm}

\usepackage[pagebackref=true,breaklinks=true,letterpaper=true,colorlinks,bookmarks=false]{hyperref}

\def\iccvPaperID{5432} 

\input{ourcommands}
\begin{document}

\title{CAPRI-Net: Learning Compact CAD Shapes with Adaptive Primitive Assembly}


\author{Fenggen Yu $^1$ \thanks{Joint first authors}
\quad
Zhiqin Chen $^1$ \footnotemark[1]
\quad
Manyi Li $^1$
\quad
Aditya Sanghi $^2$ \\
\quad
Hooman Shayani $^2$
\quad
Ali Mahdavi-Amiri $^1$
\quad
Hao Zhang $^1$ \\
$^1$Simon Fraser University \quad\quad $^2$Autodesk Research
}
\maketitle


\input{abstract}
\input{introduction}
\input{related}
\input{method}
\input{result}
\input{future}

{\small
\bibliographystyle{ieee_fullname}
\bibliography{ref}
}
\input{supp}

\end{document}

%% file: ourcommands.tex
\usepackage{color}
\definecolor{green}{rgb}{0, 0.5, 0}
\definecolor{orange}{rgb}{0.8, 0.6, 0.2}
\definecolor{red}{rgb}{1.0, 0.0, 0.0}
\definecolor{teal}{rgb}{0.0, 0.4, 0.4}
\definecolor{purple}{rgb}{0.65,0,0.65}
\definecolor{saffron}{rgb}{0.95,0.75,0.2}
\definecolor{turquoise}{rgb}{0.0,0.5,0.5}
\definecolor{black}{rgb}{0.0, 0.0, 0.0}
\definecolor{gray}{rgb}{0.5, 0.5, 0.5}

\newcommand{\rz}[1]{{\color{black}#1}}

\newcommand{\nv}[1]{\mathbf{#1}}

%% file: abstract.tex
\begin{abstract}
We introduce CAPRI-Net, a neural network for learning {\em compact\/} and {\em interpretable\/} implicit representations of 3D computer-aided design (CAD) models, in the form of {\em adaptive primitive assemblies\/}. Our network takes an input 3D shape that can be provided as a point cloud or voxel grids, and reconstructs it by a compact assembly of quadric surface primitives via constructive solid geometry (CSG) operations. The network is {\em self-supervised\/} with a reconstruction loss, leading to faithful 3D reconstructions with sharp edges and plausible CSG trees, without any ground-truth shape assemblies. While the parametric nature of CAD models does make them more predictable locally, at the shape level, \rz{there is a great deal of structural and topological variations,} which present a significant generalizability challenge to state-of-the-art neural models for 3D shapes. Our network addresses this challenge by adaptive training with respect to each test shape, with which we {\em fine-tune\/} the network that was pre-trained on a model collection. We evaluate our learning framework on both ShapeNet and ABC, the largest and most diverse CAD dataset to date, in terms of reconstruction quality, shape edges, compactness, and interpretability, to demonstrate superiority over current alternatives suitable for neural CAD reconstruction.
\end{abstract}

%% file: introduction.tex
\section{Introduction}
\label{sec:intro}

Computer-Aided Design (CAD) models are ubiquitous in engineering and manufacturing
to drive decision making 
and product evolution related to 3D shapes and geometry. With the rapid advances in AI-powered 
solutions across all relevant fields, several CAD datasets~\cite{ABC,FabWave,Fusion360} have emerged 
to support research in geometric deep learning. A common characteristic 
of CAD models is that they are composed by well-defined parametric surfaces meeting along sharp 
edges. While the parametric nature of the CAD shapes does make them more predictable locally and 
at the primitive level, at the {\em shape\/} level, there is a great deal of {\em structural\/} and {\em topological variations\/}, 
which presents a significant {\em generalizability\/} challenge to current
neural models for 3D shapes~\cite{atlasnet,imnet,OccNet,DeepSDF,bspnet,DISN,GRASS,StructureNet,wu2020pq}.
On the other hand, existing networks for primitive fitting typically focus on abstractions with simple, hence
limited, primitives~\cite{paschalidou2019superquadrics,tulsiani2017learning,3DPRNN,Im2Struct}, 
hindering reconstruction quality.

\begin{figure}[!t]
    \centering
    \includegraphics[width=0.9\linewidth]{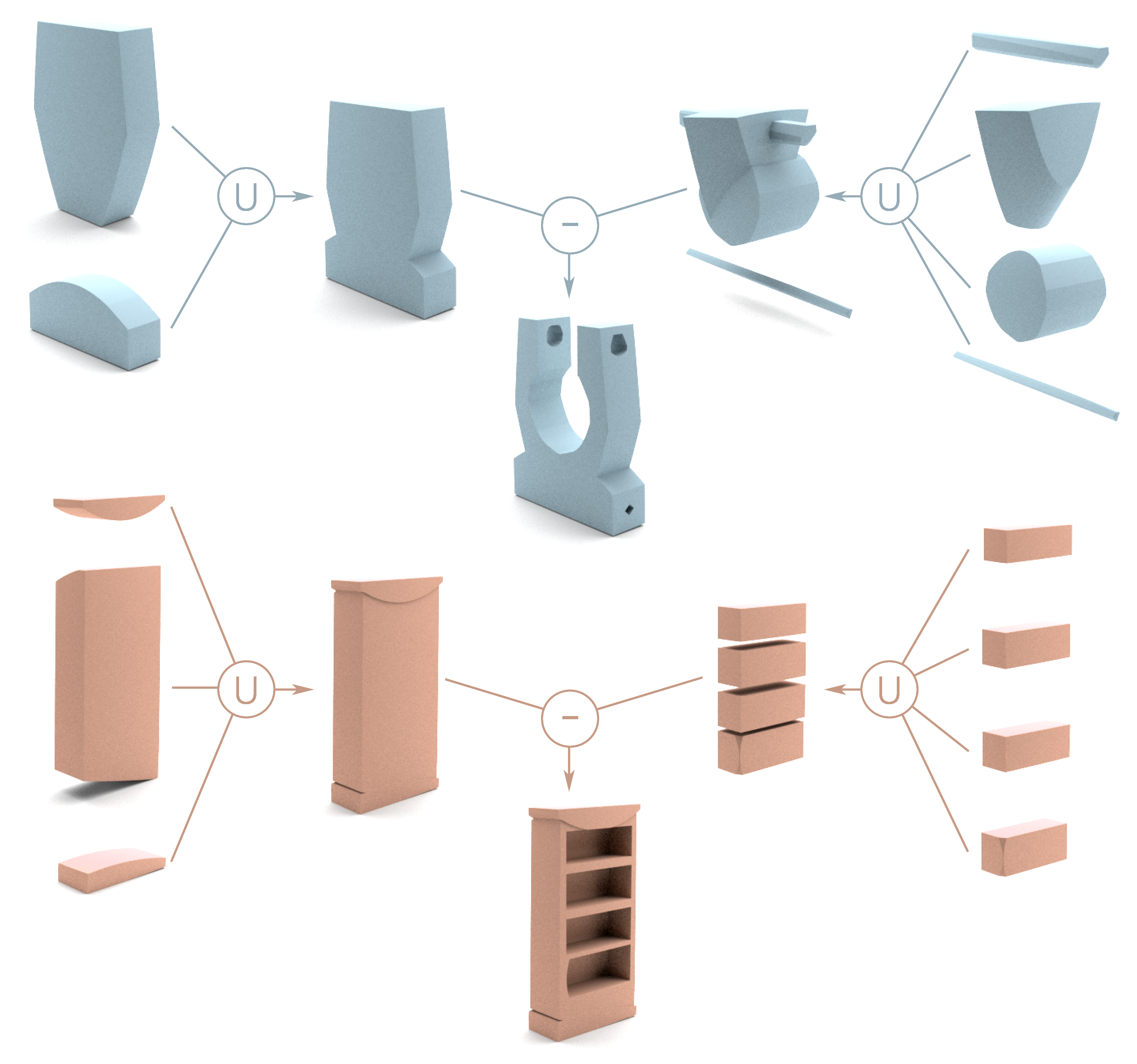}
    \caption{\rz{Our network learns compact and interpretable implicit representations of 3D CAD shapes in the form of primitive assemblies via CSG operations, without any assembly supervision.}}
    \label{fig:teaser}
\end{figure}

\begin{figure}[!t]
    \centering
    \includegraphics[width=0.99\linewidth]{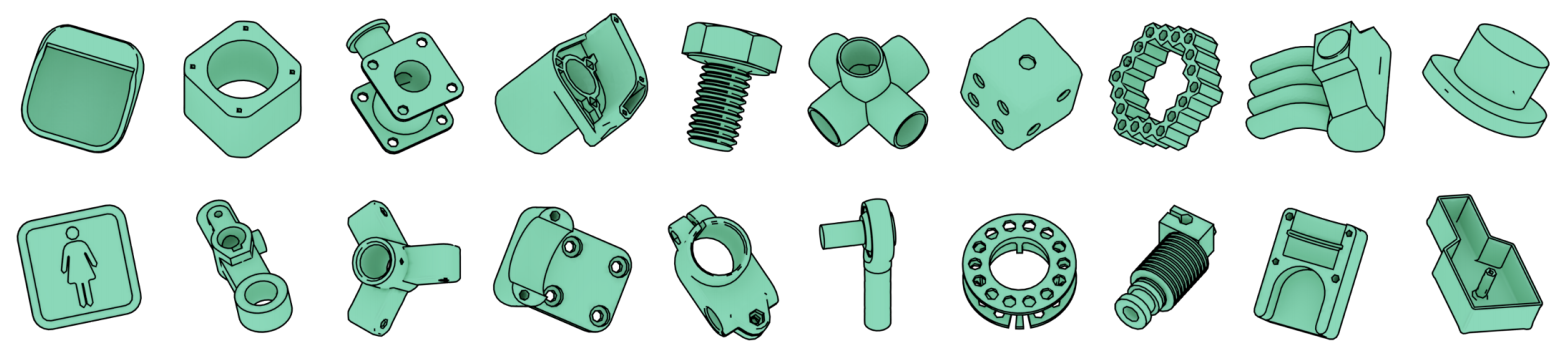}
    \caption{Random CAD shapes from the ABC dataset~\cite{ABC} to show its diversity (image taken from the original paper). }
    \label{fig:ABC}
\vspace{-10pt}
\end{figure}

In this paper, we develop a learning framework for 3D CAD shapes to address these very challenges. 
Our goal is to design a neural network which can learn a {\em compact\/} and {\em interpretable\/}
representation for CAD models, leading to high-quality 3D reconstruction, while the network generalizes well over ABC~\cite{ABC}, 
the largest and most {\em diverse\/} CAD dataset to date. Figure~\ref{fig:ABC} shows a sampler of models from ABC. This dataset
is a collection of one million CAD models covering a wide range of structural, topological, and geometric 
variations, {\em without\/} any category labels, in contrast to other prominent repositories of man-made 
shapes such as ShapeNet~\cite{chang2015shapenet} and ModelNet~\cite{3DShapeNet} \rz{which have only
limited\footnote{\rz{While the full ShapeNet dataset has 270 object categories, to the best of our knowledge, most learning
methods only work with up to 13.}} number of object categories. Hence, targeting the
ABC dataset poses a real generalizability challenge.}

\begin{figure*}
  \includegraphics[width=0.99\linewidth]{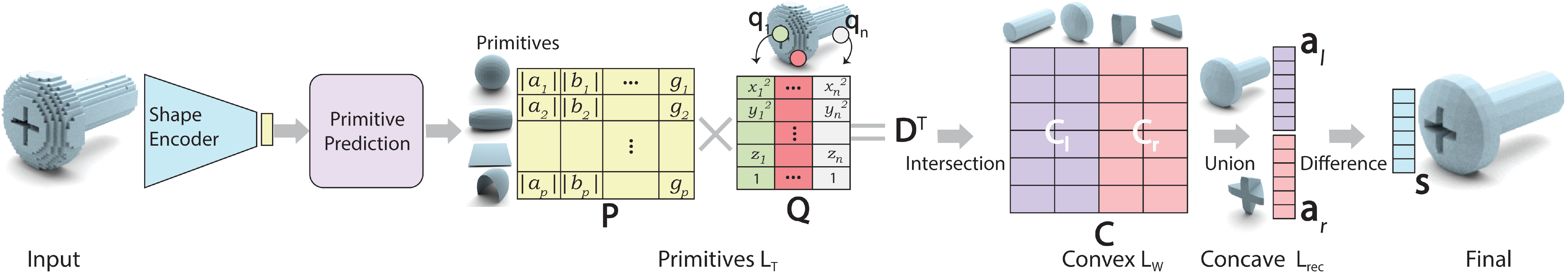}
  \caption{Overview of our network. Given \rz{an input 3D shape as a point cloud or voxels,} we first map it into a latent code using an encoder. This latent code is used to predict $p$ primitives with parameters included in $\nv{P}$. For any query point $\nv{q}_j$ packed in matrix $\nv{Q}$, we can obtain the matrix $\nv{D}$ indicating approximate signed distance from the query point to each primitive.  A selection matrix $\nv{T}$ is used to select a small set of primitives from the primitive set to group convex shapes in matrix $\nv{C}$ which indicates inside/outside values for query points w.r.t convex shapes. Then, we perform min operation on each half of $\nv{C}$ (i.e. $\nv{C}_l$ and $\nv{C}_r$) to union convex shapes into two (possibly) concave shapes and get inside/outside indication vectors $\nv{a}_{l}$ and $\nv{a}_{r}$ for left and right concave shapes. Finally, we perform a difference operation as $\nv{a}_{l} - \nv{a}_{r}$ to obtain the final point-wise inside/outside indicator $\nv{s}$. $L_\nv{T}$, $L_\nv{W}$, and $L_{rec}$ are the loss functions we define for our network.}
  \label{fig:network}
\end{figure*}

Our network takes \rz{an input 3D shape as a point cloud or voxel grids}, and reconstructs it by a compact {\em assembly\/} 
of {\em quadric surface primitives\/} via constructive solid geometry (CSG) operations including intersection, union, and difference.
Specifically, the learned quadrics are assembled by a series of binary selection matrices. These 
matrices intersect the quadrics to form convex parts, with union operations to follow to obtain possibly 
concave shapes, and finally a difference operation naturally models holes present in high-genus models, 
which are frequently encountered in CAD.

The architecture of our network, shown in Figure~\ref{fig:network}, is inspired by BSP-Net~\cite{bspnet}.
At the high level, it is a coordinate-based network trained with an occupancy loss reflecting the 
reconstruction error; we also add a novel loss to accommodate the difference operation.
The reconstruction is performed in a latent space that is obtained by an encoder
applied to the input shape. The other learnable parameters of the network include matrices that define
the parameters of the quadric surfaces and the CSG operations, respectively, as well as MLP weights
which map the latent code to the primitive parameter matrix. The resulting reconstruction is in the
form of a CSG assembly, while the network training does not require any ground-truth shape 
assemblies --- the network is {\em self-supervised\/} with the reconstruction loss.

Due to \rz{the significant variations among CAD shapes} in ABC, we found that our 
network, when trained on a model collection only, does not generalize well. In fact, none of the 
existing reconstruction networks we tested, including IM-Net~\cite{imnet}, 
OccNet~\cite{OccNet}, DISN~\cite{DISN}, BSP-Net~\cite{bspnet}, generalized well on the ABC 
dataset. We tackle this issue by further {\em fine-tuning\/} the network, that is pre-trained on
a training set, to a test CAD shape, so that the resulting network is {\em adaptive\/} to the test
shape. Both pre-training and fine-tuning are performed using the same network architecture,
shown in Figure~\ref{fig:network}, except that the encoder is not re-trained during fine-tuning
for efficiency.
We coin our learning framework CAPRI-Net, as it is trained to produce {\em Compact\/} and 
{\em Adaptive\/} {\em PRI\/}mitive assemblies for CAD shapes.

We evaluate CAPRI-Net on both ABC and ShapeNet, in terms of reconstruction quality,
compactness as measured by primitive counts, and interpretability as examined by how
natural the recovered primitive assemblies are. Qualitative and quantitative comparisons 
are made to BSP-Net and UCSG~\cite{kania2020ucsg}, a recent unsupervised method for 
inverse CSG modeling --- both are representative of state-of-the-art learning
methods suitable for CAD shapes. The results demonstrate superiority of CAPRI-Net on
all fronts.

\rz{The primary application of CAPRI-Net is 3D shape reconstruction, where the network
is trained to produce a novel implicit field that is unlike those obtained by classical 
computer graphics methods, e.g.,~\cite{carr2001,hoppe1992}, and those learned by recent neural implicit
models, e.g., IM-Net~\cite{imnet}, OccNet~\cite{OccNet}, and SIREN~\cite{sitzmann2020implicit}, etc.
Our implicit representation is {\em structured\/}, as a compact assembly, and leads to quality
reconstruction of 3D CAD models.}


%% file: related.tex
\section{Related Work}
\label{sec:RW}

We cover prior methods most closely related to our work, including classical and
emerging techniques for geometric primitive fitting, as well as recent approaches for learning implicit
and structured 3D representations.

\vspace{-8pt}

\paragraph{Primitive detection and fitting.}
%

There  has  been  extensive  work  in computer graphics and computer-aided geometric design on primitive fitting. Given a  raw 3D object, different algorithms such as RANSAC  \cite{fischler1981random} and Hough Transform  \cite{hough1959machine}  have been used to detect primitives \cite{kaiser2019survey}. RANSAC based methods have been used in \cite{schnabel2007efficient, li2011globfit}, to detect multiple primitives in dense point clouds. Methods such as \cite{du2018inversecsg, friedrich2019optimizing} have then extracted a CSG tree from the raw input.   Hough Transform has been used to detect planes \cite{borrmann20113d} and cylinders \cite {rabbani2005efficient} in point clouds. However, these methods usually do not generalize  as they require different hyper parameters per shape. Recently, neural net based algorithms have been used to detect and fit primitives on pointclouds \cite{li2019supervised, sharma2020parsenet}. SPFN \cite{li2019supervised} uses supervised learning to first detect primitive types and  then estimate the parameter of the primitives. ParseNet \cite{sharma2020parsenet} extends SPFN by in-cooperating spline patches and using differentiable metric-learning segmentation. Our method differs in requiring {\em no supervision\/}, while able to reconstruct a compact primitive assembly. 

\vspace{-8pt}

\paragraph{Neural implicit shape representations.}
Neural implicit surface representation \cite{imnet, OccNet, DeepSDF} have gained immense popularity because of the ability to generate complex, high spatial resolution 3D shapes  while using a small memory footprint during training. This representation has been used in 3D domain for part understanding \cite{chen2019bae_net, deng2019nasa, wu2020pq}, unsupervised single view reconstruction  \cite{jiang2020sdfdiff, tulsiani2017multi, niemeyer2020differentiable, liu2019learning} and scene or object completion \cite{peng2020convolutional, chibane2020implicit}.   More recently, neural implicit representations have also been extended to 2D images and videos for application in single scene completion \cite{mildenhall2020nerf, tancik2020fourier, sitzmann2020implicit}, image generative models \cite{skorokhodov2020adversarial, anokhin2020image, dupont2021generative}, super resolution \cite{chen2020learning} and dynamic scenes \cite{li2020neural, park2020deformable, xian2020space}.  In our work, we extend  the use of neural implicit surface representation to CAD models and show how our method can overcome some of the challenges presented by CAD designs. 

\vspace{-8pt}

\paragraph{Structural representations.}

Structure-aware 3D representations contain two major components:  atomic geometry entities and a structure pattern by which these atomic entities are combined \cite{egstar2020_struct}.  The atomic entities can be represented by different 3D representations, and can represent semantic meaningful parts \cite{StructureNet, GRASS, wu2020pq, sung2017complementme, wu2019sagnet, gao2019sdm} or some set of primitives  \cite{tulsiani2017learning, paschalidou2019superquadrics, deng2020cvxnet, bspnet}.  These entities can be combined to represent the complete geometry using part sets \cite{tulsiani2017learning, paschalidou2019superquadrics}, tree structure \cite{deng2020cvxnet, bspnet, sharma2018csgnet, kania2020ucsg}, relationship graphs \cite{wang2019planit, fisher2011characterizing}, hierarchical graphs \cite{StructureNet, GRASS} or programs \cite{tian2019learning}. 

In our work, we represent geometry atomic entities by using primitives defined by a constrained form of quadric implicit equation. 
Works such as Superquadrics \cite{paschalidou2019superquadrics}, have used  unconstrained explicit form of superquadratic equation but produce primitives with low interpretability as required in CAD models.  Furthermore, they cannot produce plane based intersections which produce sharp features. Our method learns the structure of combining these entities in an unsupervised manner using different CSG operations. Neural networks incooperating CSG have been explored in CSG-Net \cite{sharma2018csgnet} and UCSG-Net \cite{kania2020ucsg}. Our work differs from CSG-Net in requiring no supervision to learn the CSG tree. UCSG-Net learns the CSG tree in an unsupervised manner but uses boxes and spheres as primitives. In contrast, our work can learn more complex primitives such as cylinders while being able to reconstruct shapes at a significantly higher quality. 



Most closely related to our work is BSP-Net~\cite{bspnet}. Our work differs from BSP-Net in several significant ways: \textcircled{1} CAPRI-Net can generate several different interpretable primitives from a single quadric equation, whereas BSP-Net only uses planes. \textcircled{2}  Our network includes the difference operation which can lead to significantly less usage of primitives. \textcircled{3} Finally, we also introduce a new reconstruction loss term which can be used to encourage using the difference operation.


%% file: method.tex
\section{Methodology}
\label{sec:method}


Here, we discuss our methodology and details about our network: CAPRI-Net. 
The basic idea of CAPRI-Net is to receive a 3D shape as input and reconstruct the shape by predicting a set of primitives that are combined to make intermediate convex and concave shapes via a CSG tree.
In CAPRI-Net, primitives are represented by a quadric equation, which determines whether query point $\nv{q}_j~=~(x,y,z)$ is inside or outside that primitive.
Our selected quadric equation has the capability of producing several useful primitives including spheres, planes, or cylinders.


Voxelized inputs are fed to an encoder to produce a latent code that is passed to our primitive prediction network (see Figure \ref{fig:network}).
The output of the primitive prediction network is matrix $\nv{P}$ that holds the parameters of the predicted primitives and is used to determine the \emph{signed distance} of the query points from all predicted primitives.
These signed distances are then passed to three \emph{CSG layers} to output the occupancy value for each query point, indicating whether a point is inside or outside the shape.

\subsection{Primitive Prediction}
After obtaining a shape code with size 256 from an appropriate encoder, the code is passed to our primitive prediction network which is a multi-layer perceptron (MLP) to output primitives' parameters.
To represent primitives, we use the following quadric implicit equation:
\begin{equation}
    ax^2+by^2+cz^2+dx+ey+fz+g=0.
    \label{eq:impNeg}
\end{equation}
Therefore, our MLP outputs $\nv{P}_{p \times 7}$, where $p$ is the number of primitives represented by seven parameters $(a, b, c, d, e, f, g)$.
In fact, $\nv{P}(i,:)$, which is the $i$th row of matrix $\nv{P}$, has the parameters of the $i$th primitive.

Equation (\ref{eq:impNeg}) is capable of producing a variety of simple and complicated shapes (see Figure \ref{fig:network}) including non-convex shapes. To obtain simple yet effective primitives, we constrain the first three parameters $a, b, c$ to be {\em non-negative\/} to \rz{include frequently used convex quadric surfaces in CAD, such as planes, cylinders, ellipses, and paraboloid, but this would preclude primitives such as cones and tori.}

The quadric equation is employed to determine whether a query point is in/outside a primitive. To learn the parameters of the $p$ primitives, we sample $n$ points in the vicinity of a given shape and learn the primitive parameters to produce the same in/outside as the ground truth. 
The approximate signed distances of all points to all primitives are calculated as a matrix multiplication: $\nv{D}^T = \nv{P} \nv{Q}$. 
Given a query point $\nv{q}_j = (x_j, y_j, z_j)$, we get its signed distances to all primitives as: $\nv{D}(j,:) = \nv{P}\nv{Q}(:,j)$, where $\nv{Q}(:,j) = (x_j^2, y_j^2, z_j^2, x_j, y_j, z_j, 1)$ is the $j$th column of $\nv{Q}$. Therefore, for $n$ query points in training, signed distance matrix $\nv{D}_{n \times p}$ stores the distance between point $\nv{q}_j$ and the $i$th primitive at $\nv{D}(j,i)$.

\subsection{CSG Operations}
Constructive Solid Geometry (CSG) provides flexibility and accuracy in shape design and modeling. Therefore, being able to interpret the results in the form of a CSG tree has great benefits. In CAPRI-Net, we employ the three main CSG operations: intersection, union and difference in a fixed order to combine primitives and produce a shape. We now discuss how to incorporate these operations in our network to produce an interpretable and efficient CSG tree.


\vspace{3pt}

\noindent\textbf{Intersection.} By applying intersection operations, primitives that have been previously identified are combined to make a set of convex shapes. 
Having the signed distance matrix $\nv{D} _{n \times p}$ indicating point to primitive distances, primitives involved in forming convex shapes are selected by a selection matrix $\nv{T}_{p \times c}$. Relu activation function is first applied on $\nv{D}$ to map all the points inside the primitives to zero.
Therefore, multiplying $\text{relu}(\nv{D})$ and $\nv{T}$ provides a matrix containing point to convex distances. In fact, this matrix multiplication serves as an implicit intersection operation among primitives.
This way, only when $\nv{C}(j,i)=0$, query point $\nv{q}_j$ is inside convex $i$:
\begin{equation}
\label{eqa:hard_intersection}
    \nv{C} =  \text{relu}(\nv{D})\nv{T}
   \hspace{0.35cm}
   \begin{cases}
    0 & \text{inside,} \\
    > 0 & \text{outside.}
    \end{cases}
\end{equation}

\noindent\textbf{Union.}
After applying intersection operations on primitives and forming a set of convex shapes, the convex shapes are combined to obtain more complex and possibly concave shapes.
In CAPRI-Net, we group convex shapes into two separate shapes whose in/outside indicators are stored in $\nv{a}_{r}$ and $\nv{a}_{l}$. 
This gives the network the option to learn two shapes that can later be fed into difference layer to produce desired concavities or holes.
To do so, we first split $\nv{C}$ into two sub-matrices $\nv{C}_{l}$ and $\nv{C}_{r}$ both in size $n\times \frac{c}{2}$. $\nv{C}_{l}$ and $\nv{C}_{r}$ contain the in/outside indicators of all points to all convex shapes that are going to be combined into one shape on the left ($\nv{a}_{l}$) and another on the right ($\nv{a}_{r}$).

To compute $\nv{a}_l$ and $\nv{a}_r$, two different functions and notations are used according to our training stage. $\nv{a}^+$ and $\nv{a}^*$ respectively refer to early and later stages. Details of our multi-stage training are presented in Section \ref{subsection:training}.

We obtain $\nv{a}^*_{l}$ and $\nv{a}^*_{r}$ by applying a $\text{min}$ operation on each row of $\nv{C}_{l}$ and $\nv{C}_{r}$. This way, if point $\nv{q}_j$ is inside any of the convex shapes in $\nv{C}_l$, it will be considered as inside in $\nv{a}^*_{l}$. Assume that $\nv{C}_{l}(j,:)$ is the $j$th row of $\nv{C}_{l}$ representing in/outside indicators of point $\nv{q}_j$ with respect to convex shapes, then we define:
\begin{equation}
    \nv{a}^*_{l}(j) = \text{min}_{i < \frac{c}{2}} (\nv{C}_{l}(j,i))
    \hspace{0.35cm}
    \begin{cases}
    0 & \text{inside,} \\
    > 0 & \text{outside,}
    \end{cases}
\end{equation}
where $\nv{a}^*_{r}$ is defined similarly with $i \geq \frac{c}{2}$, and $\nv{a}^*_{l}$ and $\nv{a}^*_{r}$ are $n \times 1 $ vectors indicating whether a point is in/outside of the left/right concave shape.

Using the $\text{min}$ operation, gradients can be only back-propagated to the convex shape with the minimum value. Therefore, other convex shapes cannot be adjusted and tuned during training. To facilitate learning, we distribute gradients to all convex shapes by employing a (weighted) sum in the early stage of our multi-stage training:
\begin{equation}
    \nv{a}^+_{l}(j) = \mathscr{C}(\sum_{i < \frac{c}{2}} \nv{W}_j  \mathscr{C}(1 - \nv{C}_{l}(j,i)))
    \hspace{0.1cm}
    \begin{cases}
    1 & \approx  \text{inside,} \\
    <1 & \approx \text{outside,}
    \end{cases}
    \label{label_a_l}
\end{equation}
where $\nv{W}_j$ is a weight vector and $\mathscr{C}$ clips the values into $[0, 1]$, and $\nv{a}^+_{r}$ is defined similarly with $i \geq \frac{c}{2}$.

This function gives the chance to all elements of $\nv{C}$ to be adjusted during training.
 However, in this multiplication, small values referring to outside points can add up to a value larger than 1 which classifies an outside point as inside. Therefore, the in/outside in Equation~\ref{label_a_l} are only approximate estimates. 
 To avoid vanishing gradients, we initially set $\nv{W}_j$ to very small values (i.e., $10^{-5}$) and gradually increase $\nv{W}_j$ to be 1.
This setup helps the network converge to proper values for $\nv{a}^+$ at early stages that are finalized in $\nv{a}^*$ at later training stages. 

\vspace{3pt}

\noindent\textbf{Difference.}
Here, we introduce our novel difference operation which helps the network produce more complex and sharper objects. After we obtain $\nv{a}^*_{l}$ and $\nv{a}^*_{r}$, we perform a difference operation as below:
\begin{equation}
    \nv{s}^*(j) = \text{max}(\nv{a}^*_{l}(j), 0.2 -\nv{a}^*_{r}(j))
    \hspace{0.1cm}
    \begin{cases}
    0 & \text{inside,} \\
    > 0 & \text{outside,}
    \end{cases}
\end{equation}
\begin{equation}
    \nv{s}^+(j) = \text{min}(\nv{a}^+_{l}(j), 1-\nv{a}^+_{r}(j))
    \hspace{0.1cm}
    \begin{cases}
    1 & \approx \text{inside} \\
    < 1 & \approx \text{outside}
    \end{cases}
\end{equation}
$\nv{s}$ is used to reconstruct a shape since $\nv{s}(j)$ indicates whether query point $\nv{q}_i$ is inside or outside.

\subsection{Multi-Stage Training and Loss Functions}
\label{subsection:training}

Directly learning a \emph{binary} selection matrix $\nv{T}$ is difficult.
In addition, as already discussed, $\nv{a}^+$ can facilitate better gradient backward propagation than $\nv{a}^*$.
Therefore, we perform a multi-stage training scheme exploiting different training strategies to gradually achieve better results. 

We start training by operations and functions with better gradients, such as those in $\nv{a}^+$. Then we switch to more accurate and interpretable operations and functions, such as those in $\nv{a}^*$. In summary, at stage 0, selection matrix $\nv{T}$ is not binary and the $\text{max}$ operation is not used in $\nv{a}^+$. At stage 1, $\nv{T}$ is still not binary, but we use the $\text{max}$ operation to obtain $\nv{a}^*$. At stage 2, $\nv{T}$ becomes binary to allow deterministic selection of the right primitives.
In the following, we discuss each training stage in detail.

\vspace{3pt}

\noindent\textbf{Stage 0.} 
At this stage, we apply the following loss:
\begin{equation}
    L^+ = L^+_{rec} + L_{\nv{T}} + L_{\nv{W}},
\end{equation}
where $L_{\nv{T}}$ and $L_{\nv{W}}$ are defined similarly as in BSP-Net \cite{bspnet} by forcing each entry of $\nv{T}$ to be between 0 and 1 and each entry of $\nv{W}$ to be approximately 1:

\begin{equation}
L_{\nv{T}} = \sum_{t \in T } \text{max}(-t, 0) + \text{max}(t -1, 0),
\end{equation}

\begin{equation}
L_{\nv{W}} = \sum_{i} |\nv{W}_i -1|.
\end{equation}
$L^+_{rec}$ in CAPRI-Net differs from the usual reconstruction losses such as $L_1$ or $L_2$ that are normally used in geometric deep learning networks. 
Note that $\nv{s}$ is considered as the final output of our network to indicate whether a query point is inside or outside the shape and $\nv{g}$ holds ground truth values of these values.
Instead of defining the loss directly on $\nv{s}$ and $\nv{g}$, we use two weighted $L_2$ losses for $\nv{a}^+_{l}$ and $\nv{a}^+_{r}$ separately to make them complement each other and avoid vanishing gradients. Our $L^+_{rec}$ is defined as follows:
\begin{equation}
\begin{aligned}
    L^+_{rec} = \frac{1}{n} \sum_{j=1}^{n}[ &
    \nv{M}_{l}(j)*(\nv{g}(j) - \nv{a}^+_{l}(j))^2 +\\
     &\nv{M}_{r}(j)*((1-\nv{g}(j)) - \nv{a}^+_{r}(j))^2],
\end{aligned}
\end{equation}
where $n$ is the number of query points, and we define $\nv{M}_l(j)=\text{max}(\nv{g}(j), \mathbbm{1}( \nv{a}^+_{r}(j)\!<\!0.5) )$ and similarly define $\nv{M}_{r}(j)=\text{max}(\nv{g}(j), \mathbbm{1}(  \nv{a}^+_{l}(j)\!>\!0.5) )$. 

Function $\mathbbm{1}$ transfers Boolean values to float, while $\nv{M}_l$ and $\nv{M}_r$ adjust the losses on each side with respect to the value of the other side. 

\vspace{3pt}

\noindent\textbf{Stage 1.} At this stage, we use $\nv{a}^*$, instead of $\nv{a}^+$, to encourage the network to produce more accurate in/outside values for the left and right shapes. The loss function for this stage is given as: 
\begin{equation}
    L^* = L^*_{rec} + L_{T},
\end{equation}
where $L^*_{rec}$ is a reconstruction loss for $\nv{a}_{l}$ and $\nv{a}_{r}$ as below:

\begin{equation}
\label{eqa:loss_stage2}
    L^*_{rec} = L^*_{l} + L^*_{r},
\end{equation}

\vspace{-3pt}

\begin{equation}
\label{eq:lstarleft}
\begin{aligned}
L^*_{l} = \frac{1}{n} \sum_{j=1}^{n}&\nv{M}_{l}(j)*
[(1-\nv{g}(j))*(1- \nv{a}^*_{l}(j))+\\
&w_{l}*\nv{g}(j)*\nv{a}^*_{l}(j)], \text{and}
\end{aligned}
\end{equation}

\vspace{-3pt}

\begin{equation}
\begin{aligned}
L^*_{r} = \frac{1}{n} \sum_{j=1}^{n} &\nv{M}_{r}(j)* [\nv{g}(j)*(1 - \nv{a}^*_{r}(j))+ \\
&w_{r} * (1-\nv{g}(j))* \nv{a}^*_{r}(j)],
\end{aligned}
\end{equation}
where $\nv{M}_l(j)=\text{max}(\nv{g}(j), \mathbbm{1}( \nv{a}^*_{r}(j)\!>\!0.01) )$ and similarly $\nv{M}_{r}(j)=\text{max}(\nv{g}(j), \mathbbm{1}(  \nv{a}^*_{l}(j)\!<\!0.01) )$ and $\nv{g}$ acts like a mask. Therefore, Equation \ref{eq:lstarleft} acts similar to an $L_1$ loss where $(1-\nv{g})*(1-\nv{a}_l)$ encourages the outside points to be one and $\nv{g}*\nv{a}_l$ encourages the inside points to remain inside by receiving value zero.
$w_{l}$ and $w_{r}$ are weights to control shape decomposition structure by encouraging the left shape to cover the volume occupied by the ground truth shape and the right shape to cover a meaningful residual volume. This way an effective subtraction is obtained that is capable of producing sharp and detailed shapes with concavities and holes. Here, we set $w_{l}= 10 $ and $w_{r} = 2.5$ for all experiments. We will show effects of both weights in our ablation study.

\vspace{3pt}

\noindent\textbf{Stage 2.} In the first two stages, we use $L_{T}$ to make each entry of selection matrix $\nv{T}$ to be a float value between 0 and 1 facilitating the learning, but this $\nv{T}$ is not CSG interpretable. Therefore, we quantize $\nv{T}$ into $\nv{T}_{hard}$ with float values into binary values (i.e., $t_{hard}=(t>\eta)?1:0$), and use intersection operation with $\nv{T}$ replaced by $\nv{T}_{hard}$ (Equation~\ref{eqa:hard_intersection}) for each convex shape. Since values are small in $\nv{T}$, we set $\eta = 0.01 $ in all of our experiments. 
With the quantized matrix $\nv{T}_{hard}$, we use the same loss function in Equation~\ref{eqa:loss_stage2} as stage 1 to train the network. 



%% file: result.tex
\section{Result and Evaluation}
\label{sec:result}

In this section, we provide qualitative and quantitative results and experiments to demonstrate the effectiveness of our network. We examine the performance of CAPRI-Net on ABC dataset containing CAD shapes and also ShapeNet. We compare the performance of CAPRI-Net in shape reconstruction from voxels and point clouds with other state-of-the-art methods. We also perform ablation studies to assess the efficiency of our design.

\subsection{Data Preparation and Training Details}

\rz{We {\em randomly\/} chose, from the ABC dataset, 5,000} single-piece shapes whose normalized bounding box has sides all larger than 0.1. We discretize these shapes into $256^3$ voxels to sample 24,576 points close to the surface with their corresponding occupancy values for pre-training. 
In addition, we sample $64^3$ voxels for the input shape that is passed to the shape encoder.
For ShapeNet, we employ the dataset provided by IM-Net~\cite{imnet}, which contains $64^3$ voxelized and flood-filled 3D models from ShapeNet Core (V1) with sample points close to surface for pre-training.
Since achieving satisfactory performance by other methods on ABC also requires fine-tuning and this is time consuming (e.g., 15 minutes per shape for BSP-Net), we have selected a smaller subset of shapes as test set to evaluate the performance of fine-tuning: 50 shapes from ABC, and 20 from each category of ShapeNet (13 categories, 260 shapes in total).


%


In our experiments, we set the number of primitives as $p=1,024$ and the number of convex shapes as $c=64$. The size of our latent code for all input types is 256 and a two-layer MLP is also used to predict the parameters of the primitives from the shape latent code.

%

We first pre-train our network on a training set so as to obtain a general and initial estimate of the primitive parameters and selection matrices. During fine-tuning, conducted per test shape, we only optimize the latent code, primitive prediction network and selection matrix for each shape individually. We pre-train our network in stage 0 on our training set with 1,000 epochs, which takes $\approx9$ hours using an NVIDIA RTX 1080Ti GPU. We fine-tune our network in all stages with 12,000 iterations for each stage, taking about 3 minutes per shape overall to complete.

\subsection{Post-processing for CAD Shapes}
\label{subsec:mesh_outputting}

After obtaining the primitives $\nv{P}$ and selection matrix $\nv{T}$, we can assemble primitives into convex shapes and perform CSG operations to output a CAD mesh. 
Given a primitive selection matrix for each shape, a primitive that has been selected for some convex shapes may not be used in the formation of that convex shape if it falls outside the convex shape. Therefore, such primitives do not have an influence on the final shape and should be removed. 

To achieve this, we sample some points close to the surface of the reconstructed shape and obtain their occupancy values. Then, we remove each primitive from the list to test whether it changes the occupancy values. If after removing a primitive, the occupancy values of all points remain intact, we discard it from the primitive list.
We construct the surface of primitives that are not removed via marching cubes and perform CSG operations resulting from our network to reconstruct the final mesh. This process produces shapes with sharp edges and more regular surfaces.

\subsection{Reconstruction from Voxels}
\label{sec:3d_ae}

Given a low resolution input in voxel format (i.e. $64^3$), the task is to reconstruct a CAD mesh. First, we pre-train our entire network with our training set. We then fine-tune the shape's latent code, primitive prediction network, and the selection matrix for each shape individually.

At the fine-tuning stage, we first sample voxels whose centers are close to the shape's surface (i.e. with distance up to $1/64$). We specifically keep voxels whose occupancy values differ from their neighbors.
We then randomly sample other voxels' centers and obtain 32,768 points. Sampled points are then scaled into the range $[-0.5, 0.5]$, these points along with their occupancy values are used to supervise the fine-tuning step.

We compare CAPRI-Net with state-of-the-art networks, BSP-Net and UCSG, which output structured parametric primitives. For a fair comparison, we fine-tune these networks with the same number of iterations as well. For each shape, BSP-Net needs about $15$ minutes and UCSG about $40$ minutes to converge while CAPRI-Net only requires only about $3$ minutes. 
\rz{In supplementary material, we show results with and without fine-tuning.}

\input{Tables/3d_voxel_quan}

\begin{figure}
\centering
  \includegraphics[width=1.0\linewidth]{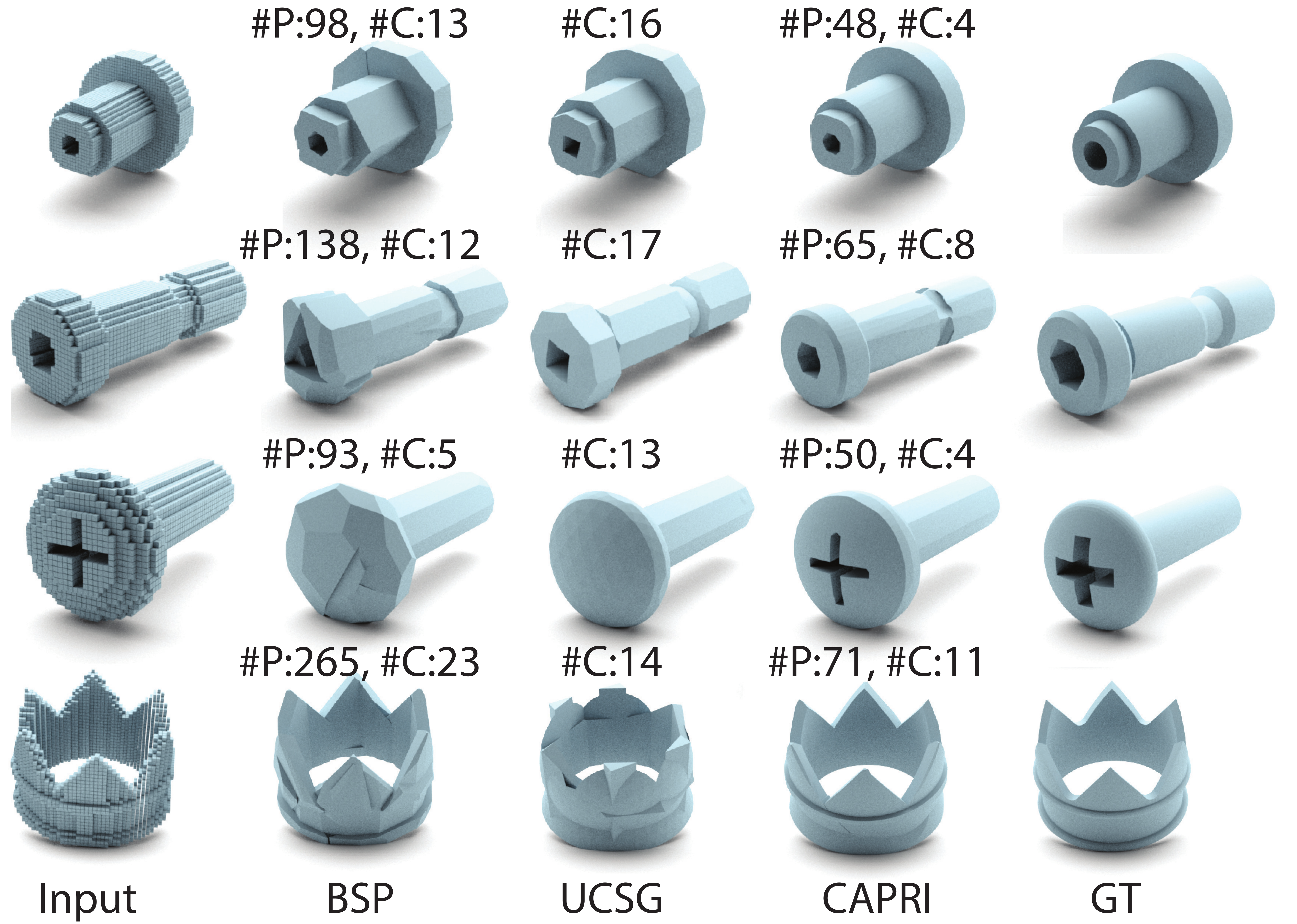}
  \caption{Visual comparisons between reconstruction results from $64^3$ voxel inputs on ABC. \rz{We also show number of surface primitives (\#P) and number of convexes (\#C) reconstructed.}}
  \label{fig:3d_abc_voxel_visual}
\end{figure}

\vspace{3pt}

\noindent\textbf{Evaluation metrics.}
Our quantitative metrics for shape reconstruction are symmetric Chamfer Distance (CD), Normal Consistency (NC, higher is better), Edge Chamfer Distance \cite{bspnet} (ECD), and Light Field Distance \cite{LFD} (LFD). For ECD, we set the threshold for normal cross products to 0.1 for extracting points close to edges. CD and ECD values provided are computed on $8k$ surface sampled points and multiplied by 1,000. For LFD, we render each shape at ten different views and measure Light Field Distances.

\vspace{3pt}

\begin{figure}
\centering
  \includegraphics[width=1\linewidth]{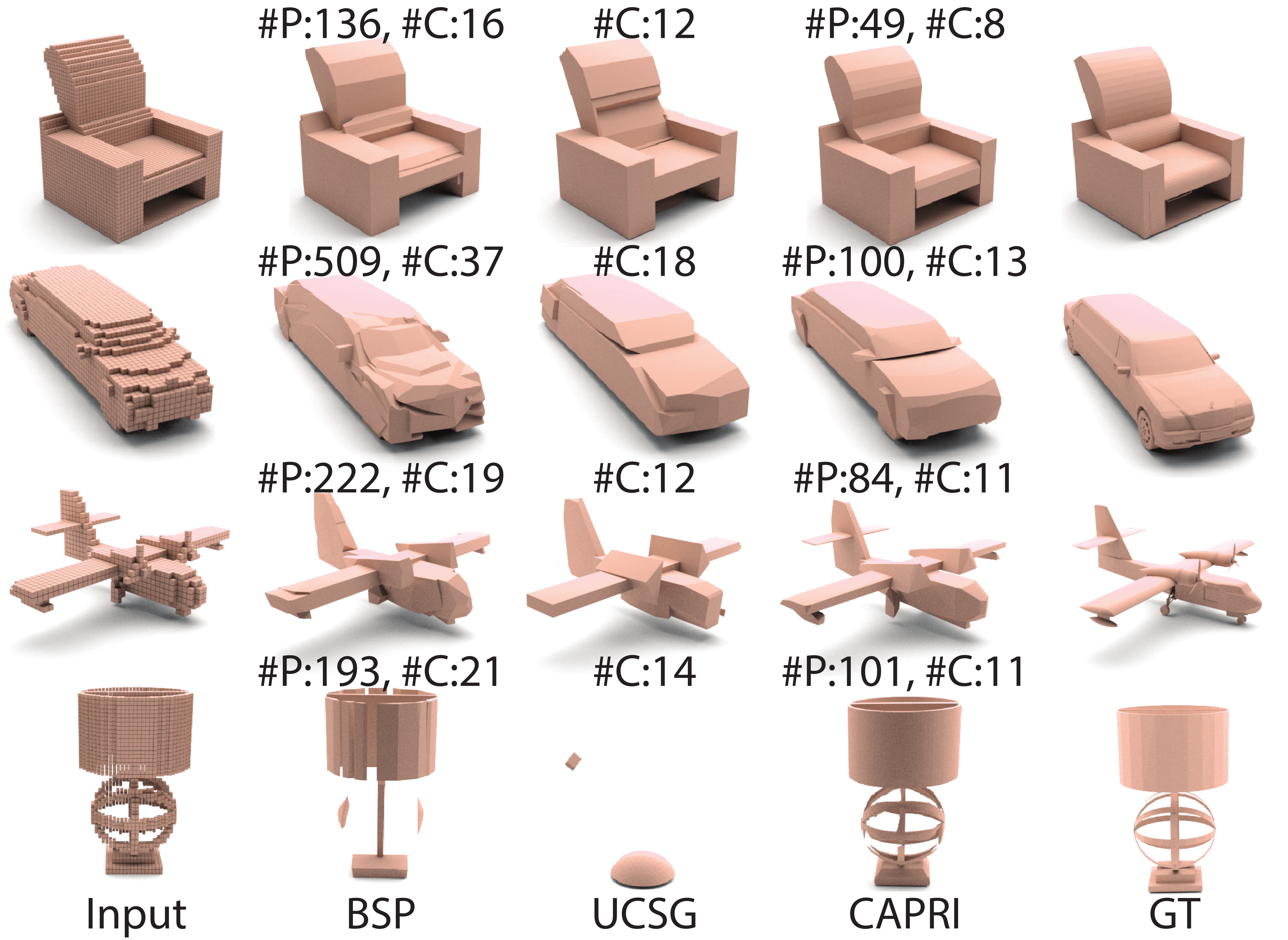}
  \caption{Some reconstruction results from voxels on ShapeNet.}
  \label{fig:3d_shapenet_voxel_visual}
\vspace{-10pt}
\end{figure}

\noindent\textbf{Evaluation and comparison.} 
We provide visual comparisons \rz{on representative examples} from the ABC dataset in Figure~\ref{fig:3d_abc_voxel_visual} and ShapeNet in Figure~\ref{fig:3d_shapenet_voxel_visual}. \rz{Our method consistently reconstructs more accurately the geometric and topological details such as holes and sharp features, owing to the richer sets of supported surface primitives and the difference operation which was not present in BSP-Net. UCSG performs well when the input can be well modeled by boxes and spheres, but not others such as the lamp in Figure~\ref{fig:3d_shapenet_voxel_visual}.}

\rz{
Quantitative comparison results are shown in Tables \ref{tab:3d_voxel_abc} and \ref{tab:3d_voxel_shapenet}. CAPRI-Net achieves the best reconstruction quality on all metrics reported, except for one instance: BSP-Net wins on NC for the ShapeNet test set.
We measure compactness of the reconstruction by counting average surface primitives and convexes the methods produce. CAPRI-Net outperforms both BSP-Net and UCSG.}


\vspace{3pt}

\noindent\textbf{CSG trees.} Our network can learn to produce a \rz{plausible} CSG tree from a given latent code without direct supervision as is shown in Figure \ref{fig:teaser}.We provide additional CSG trees and comparison results in the supplementary material. 

\vspace{3pt}

\begin{figure}
\centering
  \includegraphics[width=0.96\linewidth]{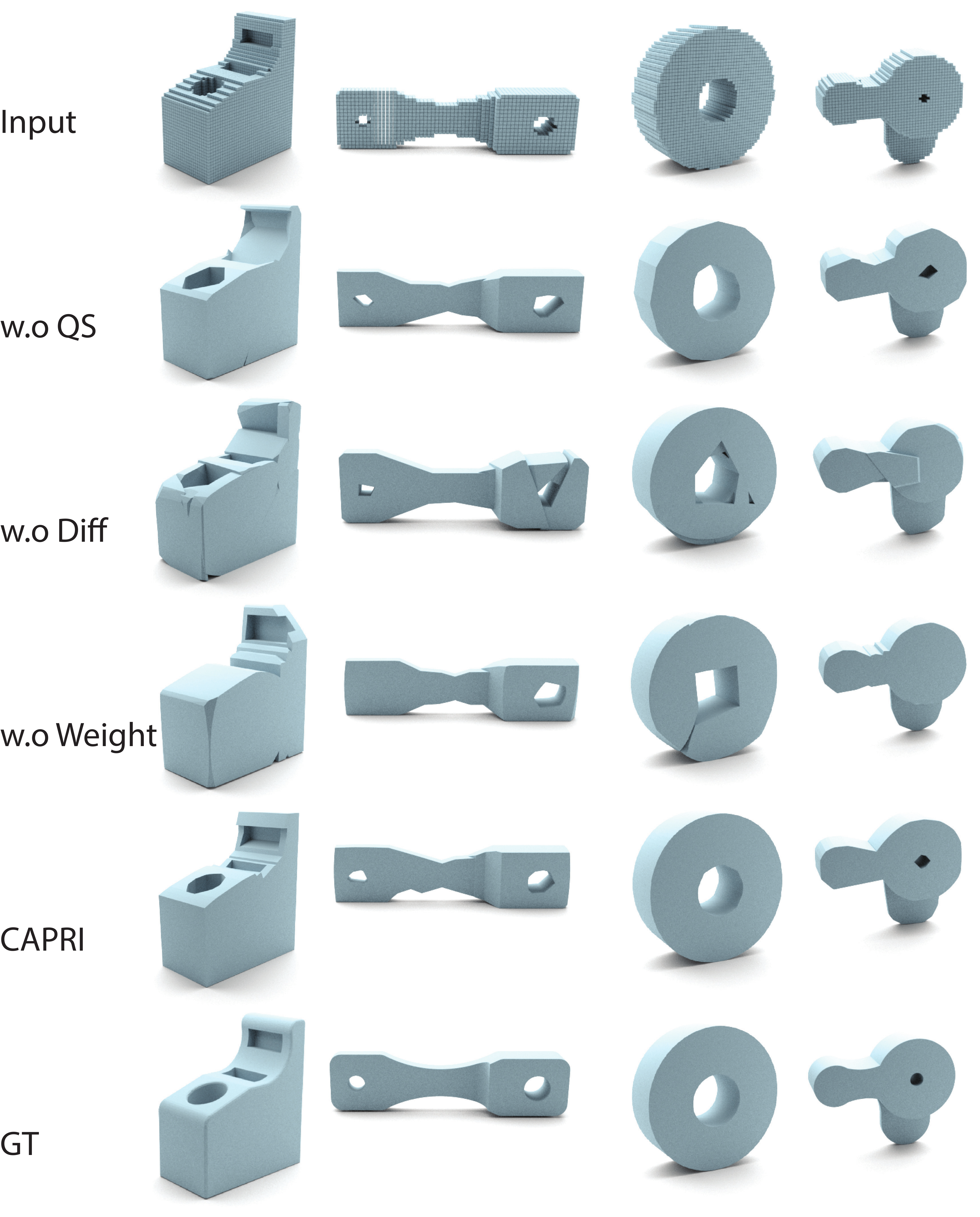}
  \caption{Visual comparison results for ablation study on ABC.}
  \label{fig:3d_voxel_ablation_visual}
\vspace{-10pt}
\end{figure}

\noindent\textbf{Ablation.} We perform an ablation study to examine the effects of three important design choices we made in CAPRI-Net: quadric surface (QS) representation, difference layer (Diff), and weighted reconstruction loss (Weight). We deactivate each of these components and make three ablation studies called: w.o QS, w.o Differ, and w.o Weight; see Figure \ref{fig:3d_voxel_ablation_visual} and Table \ref{tab:ablation_abc}. It is apparent that quadric surface representation makes our method suitable for ABC dataset by using fewer appropriate primitives (e.g., cylinders) in the reconstruction. Difference operation can also offer compactness and fewer primitives in the final reconstruction. Finally, the weighted reconstruction loss helps CAPRI-Net reproduce fine details such as small holes.

\begin{figure*}[!t]
\centering
  \includegraphics[width=.98\linewidth]{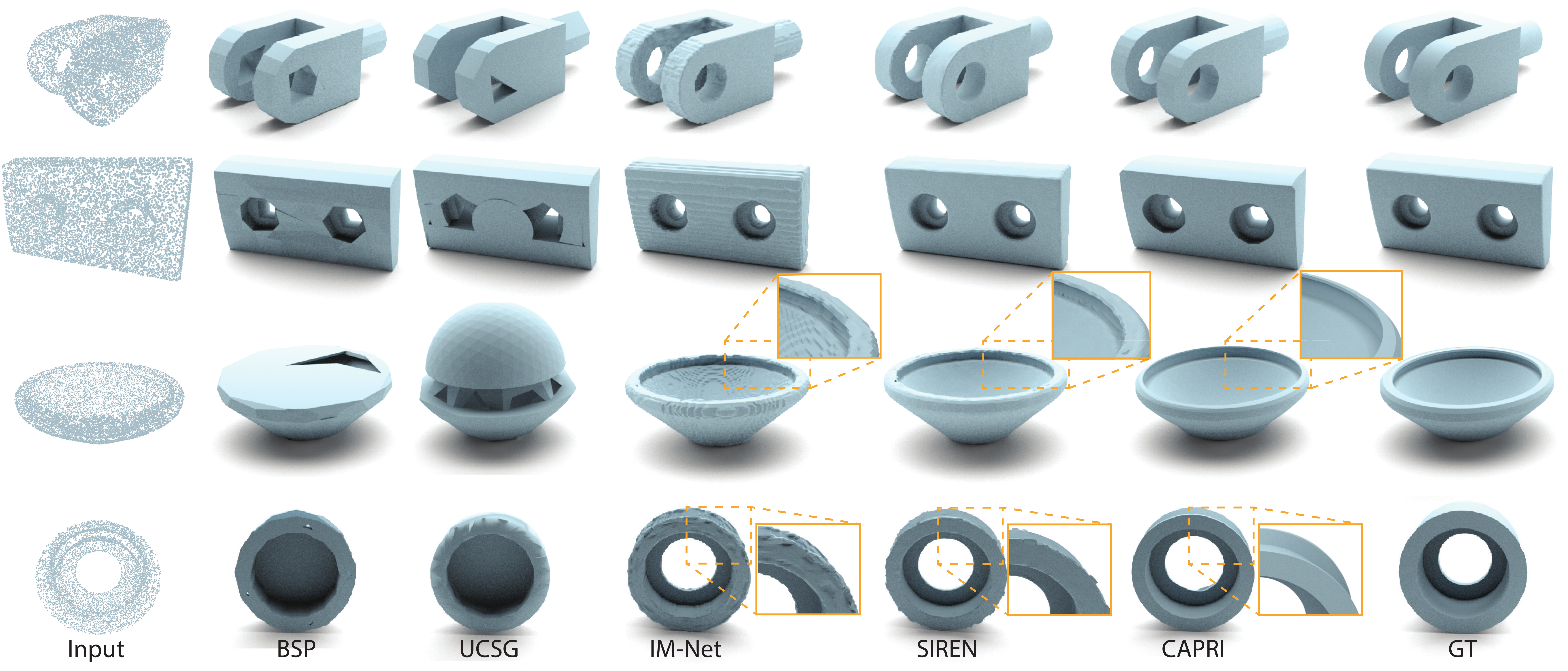}
  \caption{Visual comparisons between reconstruction results from point clouds (8,192 points) on ABC. Pay attention to the insets which show noticeable surface artifacts from IM-Net and SIREN results, \rz{both at $128^3$ resolution.}}
  \label{fig:3d_abc_pc_visual}
\vspace{-10pt}
\end{figure*}

\subsection{Reconstruction from Point Clouds}

\rz{
In our last experiment, we test reconstruction of CAD meshes from point clouds, each containing $8,192$ points with normal vectors.
During pre-training, we 
voxelize the input point clouds to $64^3$ and train a 3D convolution network as the encoder to generate the shape latent code.}

\rz{
In the fine-tuning stage, network training adapts to the original input point clouds, not the voxels. Specifically,} inspired from \cite{jiang2020local}, for each point with its normal vector, we sample $8$ points along its normal with Gaussian distribution $(\mu = 0, \sigma=1/64)$. If this point is against point normal direction, then occupancy value is 1, otherwise it is 0. This way, we can sample $65,536$ points to fine-tune the network for each shape. Similar to the fine-tuning step of voxelized inputs, we only fine-tune our latent codes, primitive prediction network and selection matrix.

\input{Tables/3d_pc_quan}

\rz{Quantitative comparisons in Table~\ref{tab:3d_pc_quan} show that our network outperforms BSP-Net and UCSG across the board. We provide additional ShapeNet comparison results in the supplementary material.
With respect to state-of-the-art, {\em unstructured\/}, non-parametric-surface learning methods such as IM-Net \cite{imnet} and SIREN \cite{sitzmann2020implicit}, CAPRI-Net produces comparable metric performances, but is slightly worse, since there is an apparent trade-off between reconstruction quality and the desire to obtain a compact primitive assembly; see supplementary material for details. In terms of visual quality however, as shown in Figure~\ref{fig:3d_abc_pc_visual}, geometric artifacts such as small bumps and pits are often present on results from IM-Net and SIREN, while the mesh surfaces produced by CAPRI-Net, are more regularized.}



%% file: Tables/3d_voxel_quan.tex
\setlength{\tabcolsep}{4pt}

\begin{table}[t!]
\centering

 \begin{tabular}{|l| r| r| r|}
     \hline
 Methods & BSP-Net & UCSG & Ours\\
    \hline \hline
CD $\downarrow$        & 0.266 & 0.182 & \textbf{0.148} \\
NC $\uparrow$ & 0.901 & 0.891 & \textbf{0.908} \\
ECD $\downarrow$       & 5.661 & 5.892 & \textbf{4.267}\\
LFD $\downarrow$       & 1,149.7 & 1,172.8 & \textbf{750.3} \\
\#Primitives (\#P) $\downarrow$ & 141.68 & - & \textbf{48.96} \\
\#Convexes (\#C) $\downarrow$    & 12.72 & 12.72 & \textbf{5.46}\\
  \hline
\end{tabular}
 \caption{\rz{Comparing 3D shape reconstruction from voxels on ABC. UCSG does not produce surface primitives, only boxes and spheres, which are convex --- we count them only under convexes. Note that the BSP-Net and UCSG averages happen to be identical.}}
 
\label{tab:3d_voxel_abc}
\end{table}

\begin{table}[t!]
\centering

 \begin{tabular}{|l| r| r| r|}
    \hline
 Methods & BSP-Net & UCSG & Ours\\
    \hline \hline
CD  $\downarrow$ & 0.239 & 0.681 & \textbf{0.201} \\
NC $\uparrow$ & \textbf{0.867} & 0.835 & 0.850 \\
ECD $\downarrow$ & 2.068 & 4.149 & \textbf{2.062}\\
LFD $\downarrow$ & 2,189.8 & 3,489.2 & \textbf{1,906.3} \\
\#Primitives (\#P) $\downarrow$ & 211.75 & - & \textbf{62.21} \\
\#Convexes (\#C) $\downarrow$ & 18.81 & 12.81 & \textbf{7.70}\\
  \hline
\end{tabular}
  \caption{\rz{Comparing 3D reconstruction from voxels on ShapeNet.} }
\label{tab:3d_voxel_shapenet}
\end{table}

\begin{table}[t!]
\centering
\vspace{2pt}
 \begin{tabular}{|l| r| r| r| r| }
    \hline
 Methods & w.o QS & w.o Diff & w.o Weight & Ours\\
    \hline \hline
CD  $\downarrow$ & 0.34 & 0.29 & 0.26 & \textbf{0.15} \\
NC $\uparrow$ & 0.89 & 0.86 & 0.89 & \textbf{0.91} \\
ECD $\downarrow$ & 5.290 & 4.408 & 5.315 & \textbf{4.267} \\
LFD $\downarrow$ & 1,477.3 & 1,323.8 & 1,047.1 & \textbf{750.3} \\
\#P $\downarrow$ & 71.98 & 77.30 & 58.44 & \textbf{48.96} \\
\#C $\downarrow$ & \textbf{5.06} & 10.42 & 5.30 & 5.46 \\
  \hline
\end{tabular}
\caption{Ablation study on ABC; see texts for descriptions.}

\label{tab:ablation_abc}
\vspace{-10pt}
\end{table}

%% file: Tables/3d_pc_quan.tex
\setlength{\tabcolsep}{3pt}

\begin{table}[t!]
\centering

\begin{tabular}{|l| r| r| r| r| r| r|}
    \hline
 Methods & CD $\downarrow$ & NC $\uparrow$& ECD $\downarrow$& LFD $\downarrow$ & \#P $\downarrow$ & \#C $\downarrow$\\
    \hline \hline
UCSG         & 1.43 & 0.88 & 11.56 & 2428.5 & - & 12.92\\
BSP-Net      & 0.19 & \bf 0.94 & 2.91 & 621.7 & 180.80 & 14.94 \\
Ours         & \bf 0.09 & \bf 0.94 & \bf 2.54 & \bf 500.5 & \bf 70.26 & \textbf{5.30} \\
  \hline
\end{tabular}
\caption{Comparing 3D point cloud reconstruction on ABC. }

\label{tab:3d_pc_quan}

\vspace{-10pt}

\end{table}

%% file: future.tex
\section{Discussion, limitation, and future work}
\label{sec:future}

In recent years, there has been a large volume of works which target ShapeNet~\cite{chang2015shapenet} 
for the development of neural 3D shape representations. We are not aware of similar 
efforts devoted to CAD models which possess {\em richer\/} geometric and topological variations, but 
lack strong structural predictability tied to a limited number of object categories. Our network, CAPRI-Net, fills
this gap as it targets the ABC dataset~\cite{ABC} whose CAD models exhibit these very characteristics. 
As our results demonstrate, our reconstruction network outperfoms state-of-the-art alternatives on 
both ABC {\em and\/} ShapeNet.
Yet, CAPRI-Net only represents an early attempt at learning primitive assemblies for CAD models
and beyond, since it is still limited on several fronts.

First, our network follows a {\em fixed\/} assembly order, i.e., intersection followed by union and then a single
difference operation. As such, not all assemblies, e.g., a nested difference, can be represented. Second,
despite the compactness exhibited by the recovered assemblies, CAPRI-Net does not have a network loss to 
enforce minimality of the CSG trees. Consistent with the minimum description length principle~\cite{MDLbook},
devising such a loss could benefit many tasks beyond our problem domain. 
\rz{Related to this, incorporating CSG operations into the network tends to cause gradient back-propagation issues.}
Third, our current fine-tuning does not work well on single-view images as input for the ABC CAD models.
Last but not the least, our current approach does not take full advantage of the {\em local\/} regularity of 
CAD models due to their parametric nature.

In addition to addressing the above limitations, we would also like to extend CAPRI-Net into a fully generative 
model for CAD design. Conditioning the generator on design or hand-drawn sketches is also a promising 
direction considering its application potential. Finally, learning {\em functionality\/} of CAD
models is also an intriguing topic to explore.

%% file: supp.tex
\vspace{10pt}
\appendix
\section{Supplementary material}

\subsection{Reconstruction from Voxels}

\begin{figure*}
  \includegraphics[width=1\linewidth]{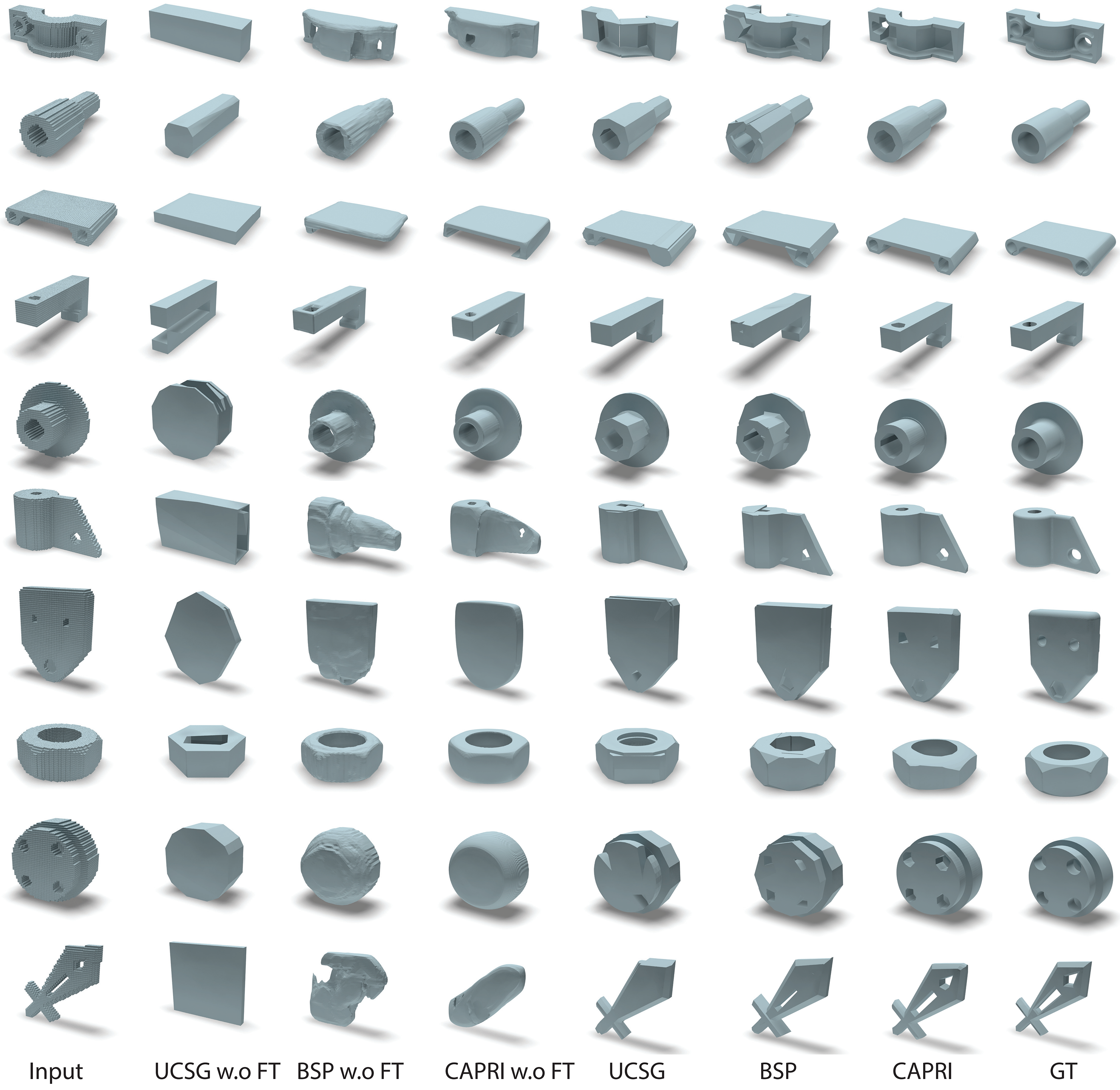}
  \caption{Visual comparisons between reconstruction results from $64^3$ voxel inputs on ABC. We also show results before fine-tuning.}
  \label{fig:abc_voxel}
\end{figure*}

\begin{figure*}
  \includegraphics[width=1\linewidth]{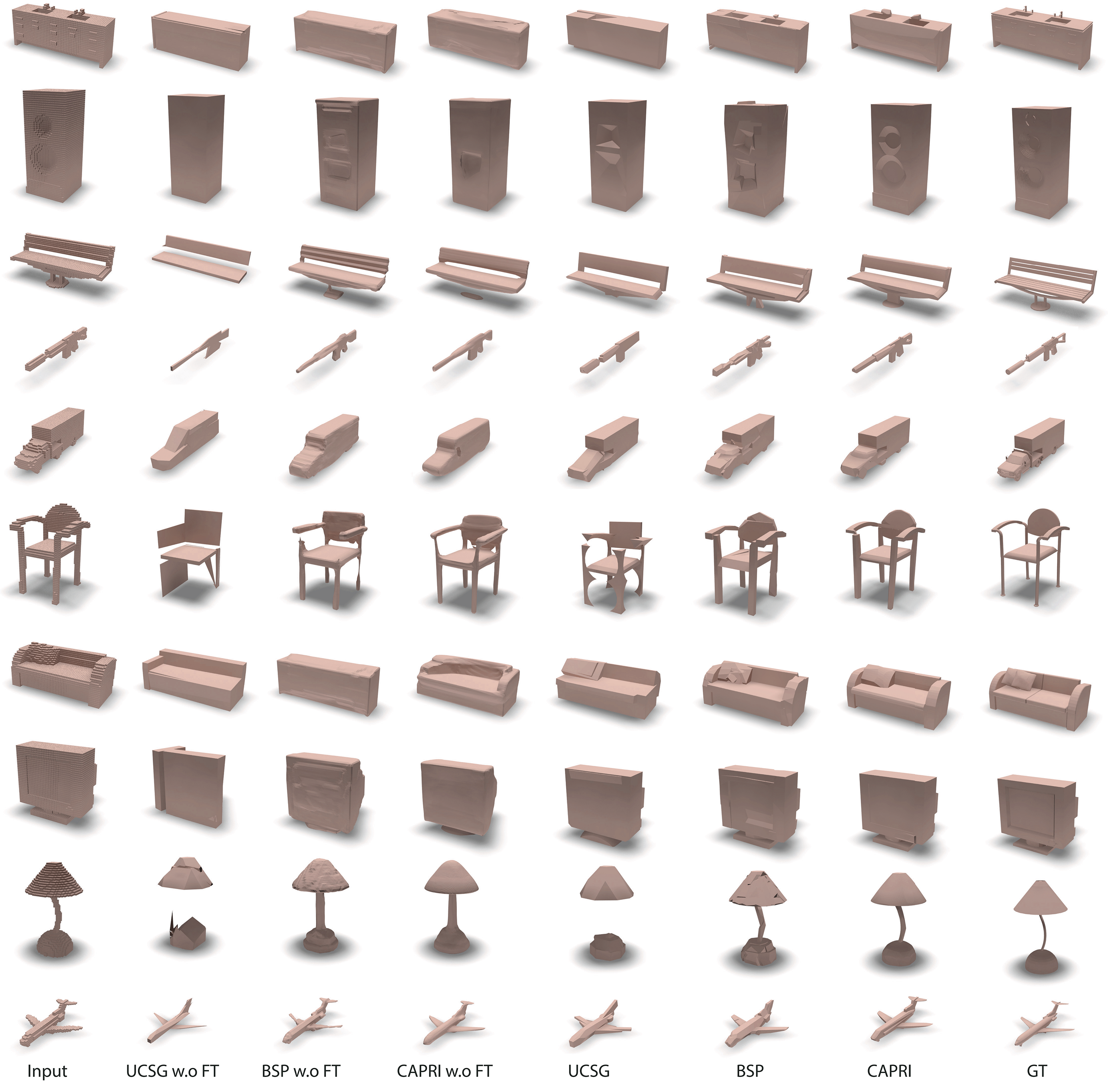}
  \caption{Visual comparisons between some reconstruction results from $64^3$ voxel inputs on ShapeNet.}
  \label{fig:shapenet_voxel}
\end{figure*}

We present more qualitative and quantitative results on shape reconstruction from voxels in Figures \ref{fig:abc_voxel} and \ref{fig:shapenet_voxel} and Tables~\ref{tab:voxel_abc} and \ref{tab:voxel_shapenet}. We also show some results without fine-tuning from BSP-Net~\cite{bspnet}, UCSG~\cite{kania2020ucsg}, and our method. These results are obtained by testing with the pre-trained networks, and we call them w.o FT. The results show that after fine-tuning, we obtain better output shapes that are more consistent with the input and attain more details.
\input{Tables/voxel}

\subsection{Reconstruction from Point Clouds}

\begin{figure*}
  \includegraphics[width=1\linewidth]{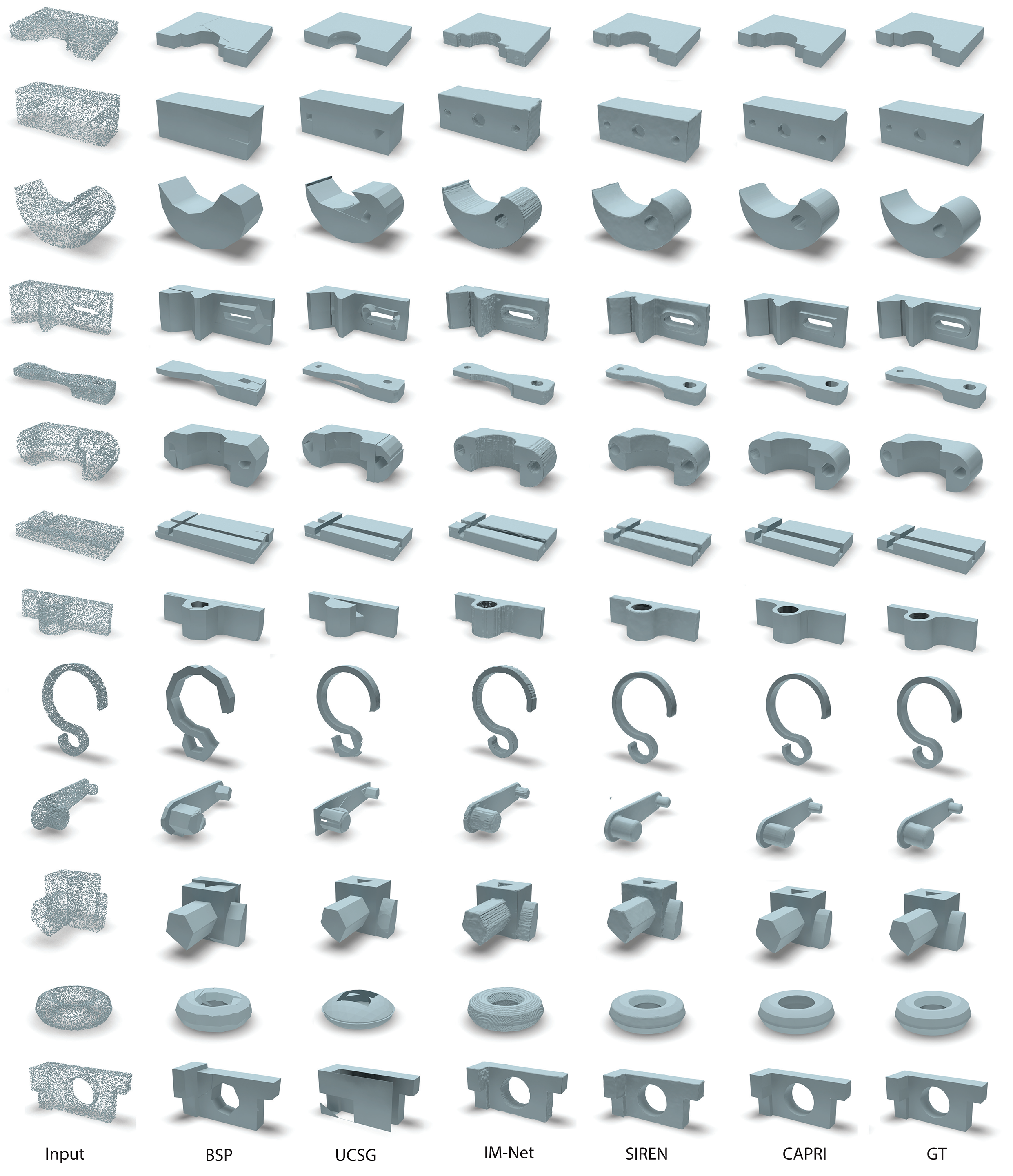}
  \caption{Visual comparisons between reconstruction results from point clouds (8,192 points) on ABC. Pay attention to the surface artifacts from IM-Net and SIREN results, both at $128^3$ resolution.}
  \label{fig:abc_pc}
\end{figure*}

\begin{figure*}
  \includegraphics[width=1\linewidth]{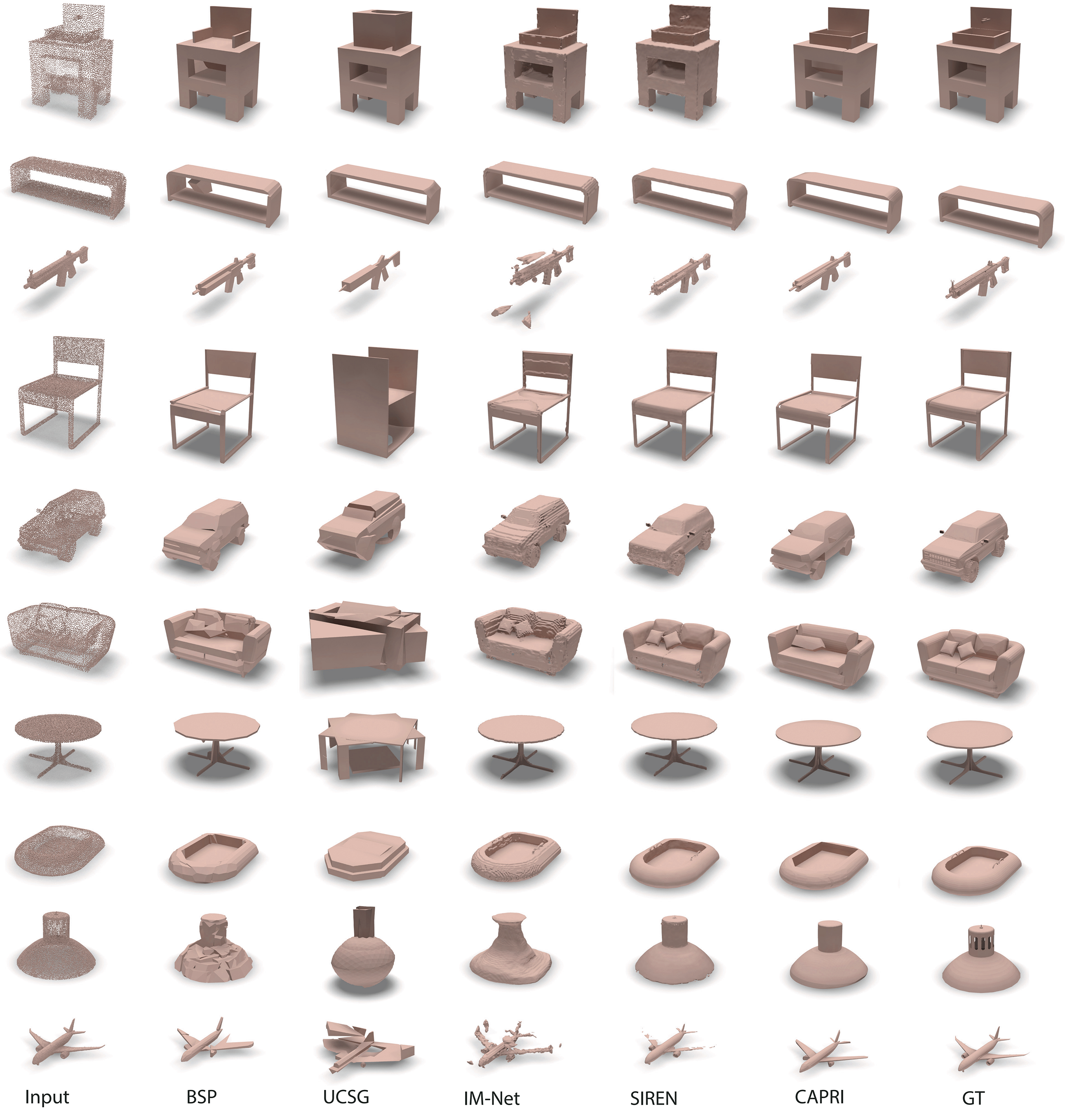}
  \caption{Visual comparisons between reconstruction results from point clouds (8,192 points) on ShapeNet.}
  \label{fig:shapenet_pc}
\end{figure*}

Tables~\ref{tab:pc_abc} and~\ref{tab:pc_shapenet} show quantitative results on ABC dataset and ShapeNet. Compared with the state-of-the-art, {\em unstructured\/}, non-parametric-surface learning methods such as IM-Net \cite{imnet} and SIREN \cite{sitzmann2020implicit}, CAPRI-Net produces comparable numeric results, but it is slightly worse, due to the trade-off between the reconstruction quality and the desire to obtain a compact primitive assembly. However, in terms of visual quality, as shown in Figure~\ref{fig:abc_pc} and Figure~\ref{fig:shapenet_pc}, the results from IM-Net and SIREN often possess geometric artifacts such as small bumps and pits, while the mesh surfaces produced by CAPRI-Net are smoother and more regular. 

\input{Tables/point}

\subsection{CSG Trees}
\begin{figure*}
  \includegraphics[width=1\linewidth]{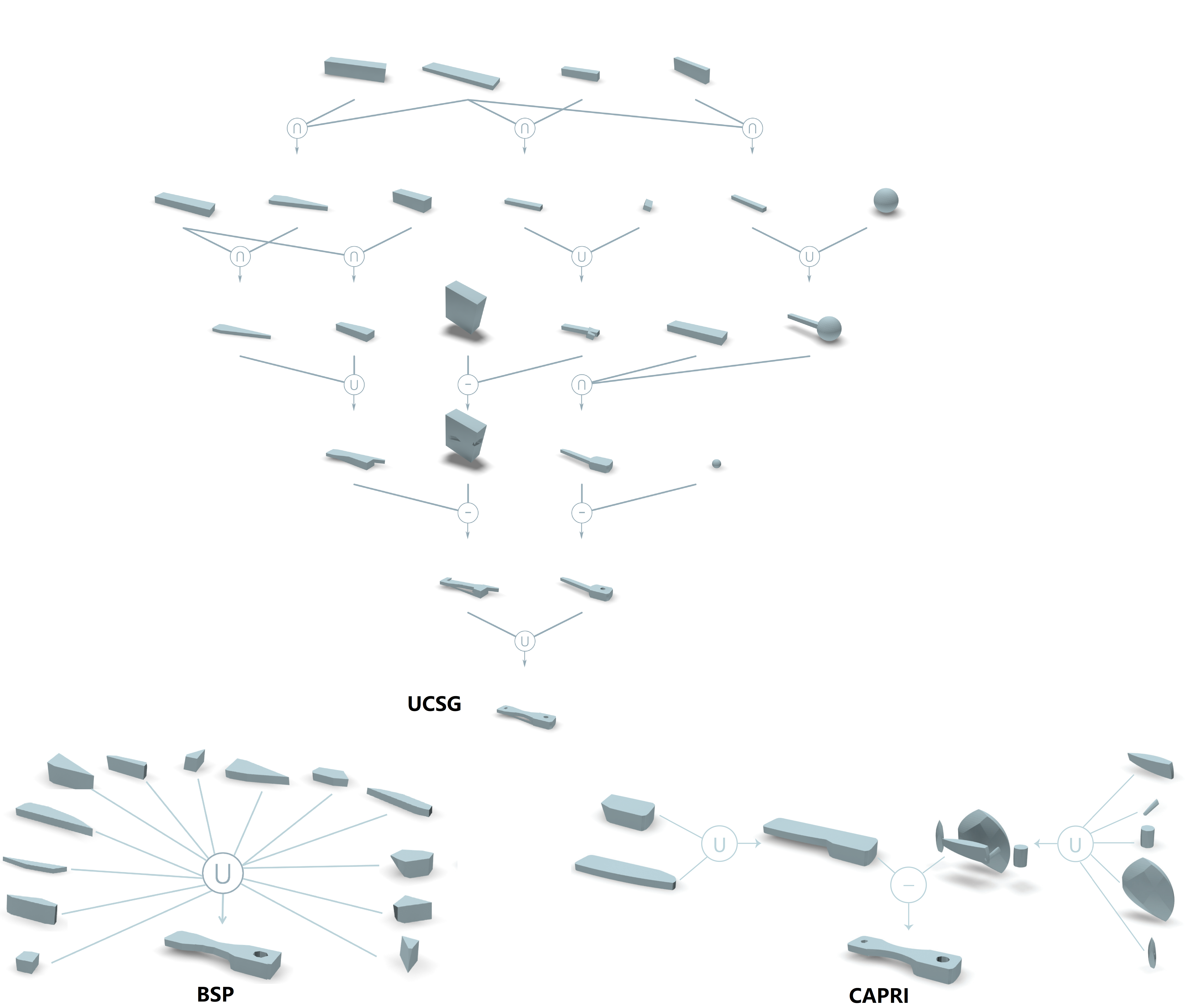}
  \caption{An example of CSG tree visual comparison on ABC.}
  \label{fig:abc_tree_1}
\end{figure*}

\begin{figure*}
  \includegraphics[width=1\linewidth]{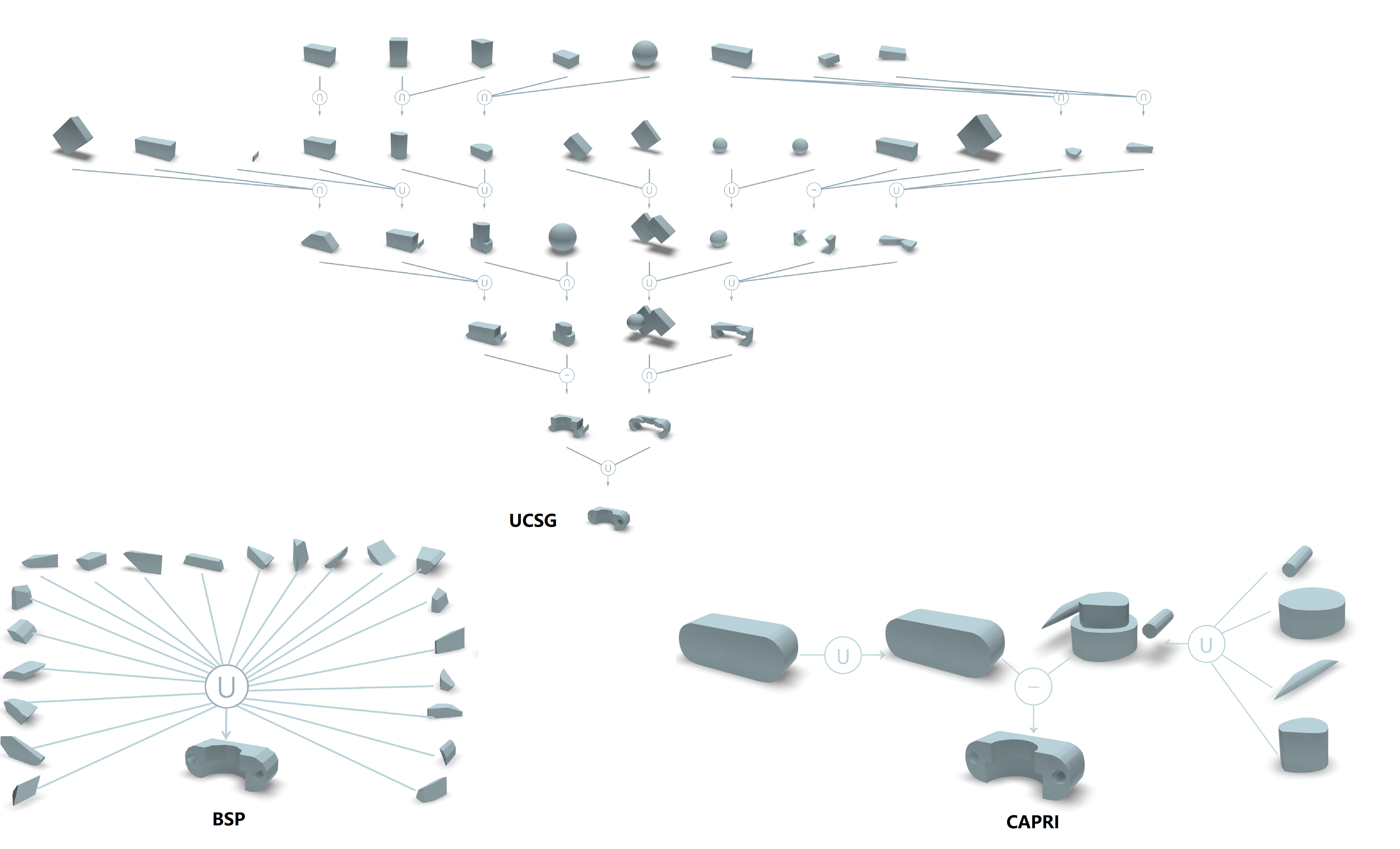}
  \caption{An example of CSG tree visual comparison on ABC.}
  \label{fig:abc_tree_2}
\end{figure*}

\begin{figure*}
  \includegraphics[width=1\linewidth]{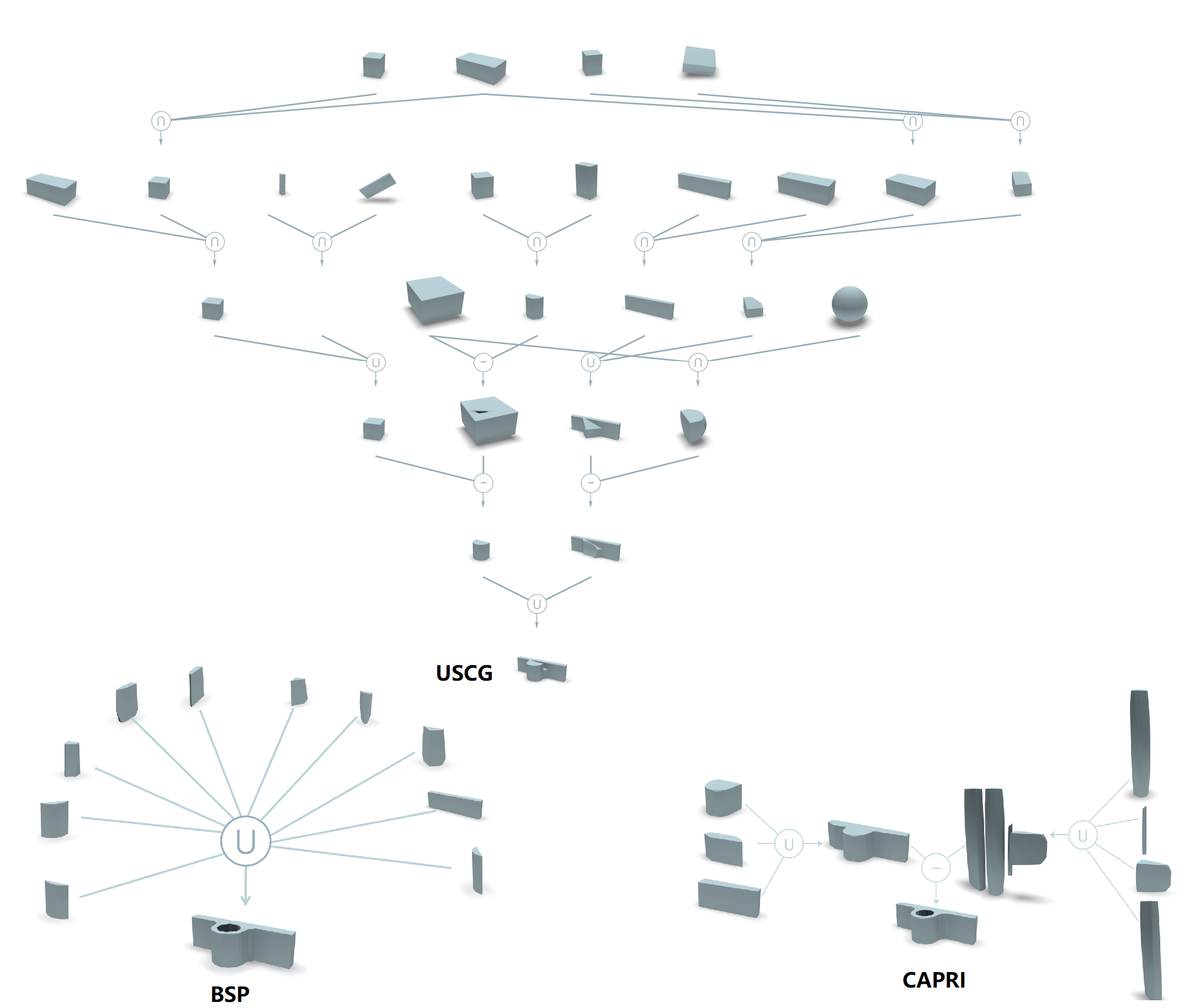}
  \caption{An example of CSG tree visual comparison on ABC.}
  \label{fig:abc_tree_3}
\end{figure*}

\begin{figure*}
  \includegraphics[width=1\linewidth]{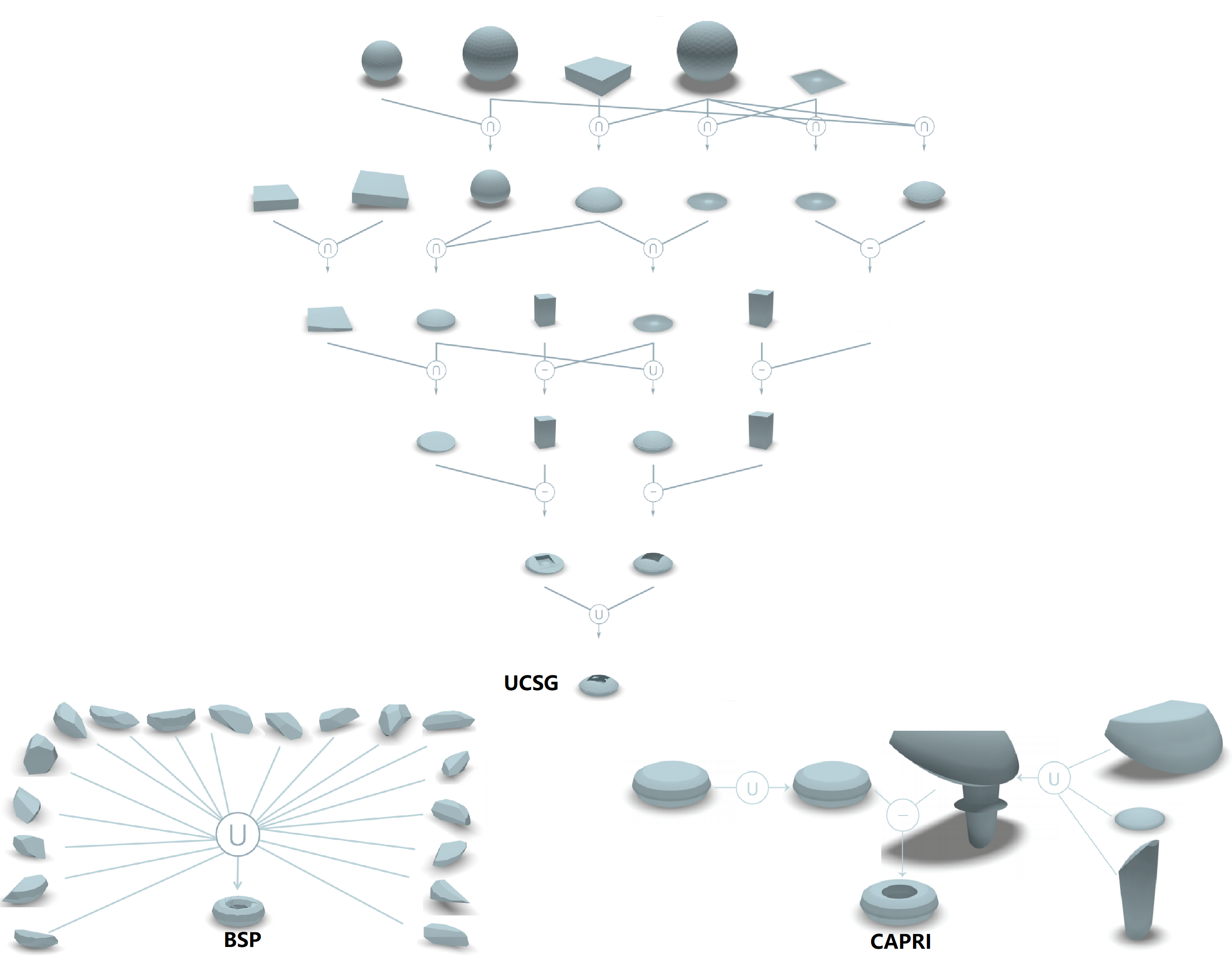}
  \caption{An example of CSG tree visual comparison on ABC.}
  \label{fig:abc_tree_4}
\end{figure*}

\begin{figure*}
  \includegraphics[width=1\linewidth]{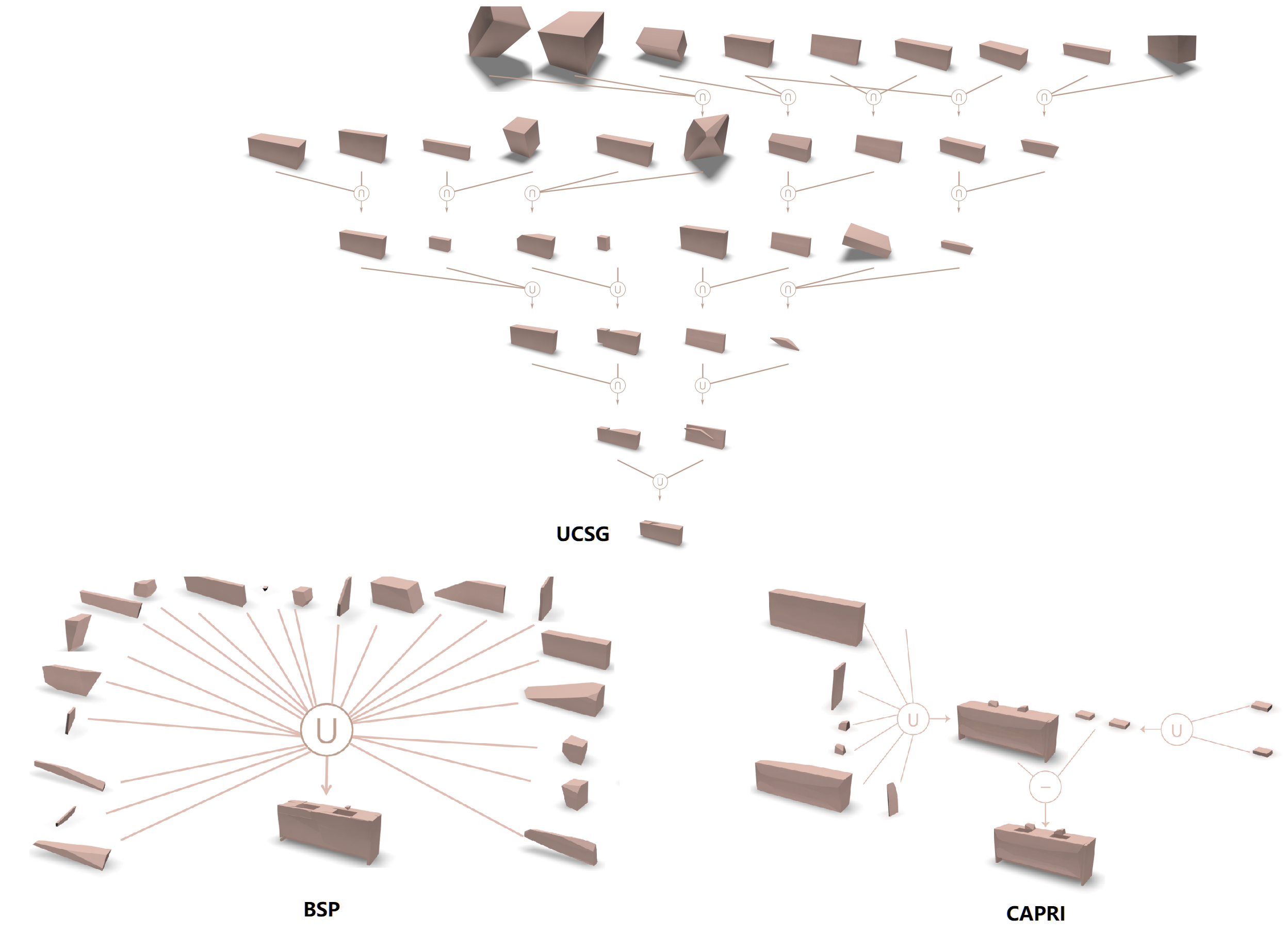}
  \caption{An example of CSG tree visual comparison on ShapeNet.}
  \label{fig:shapenet_tree_1}
\end{figure*}

\begin{figure*}
  \includegraphics[width=1\linewidth]{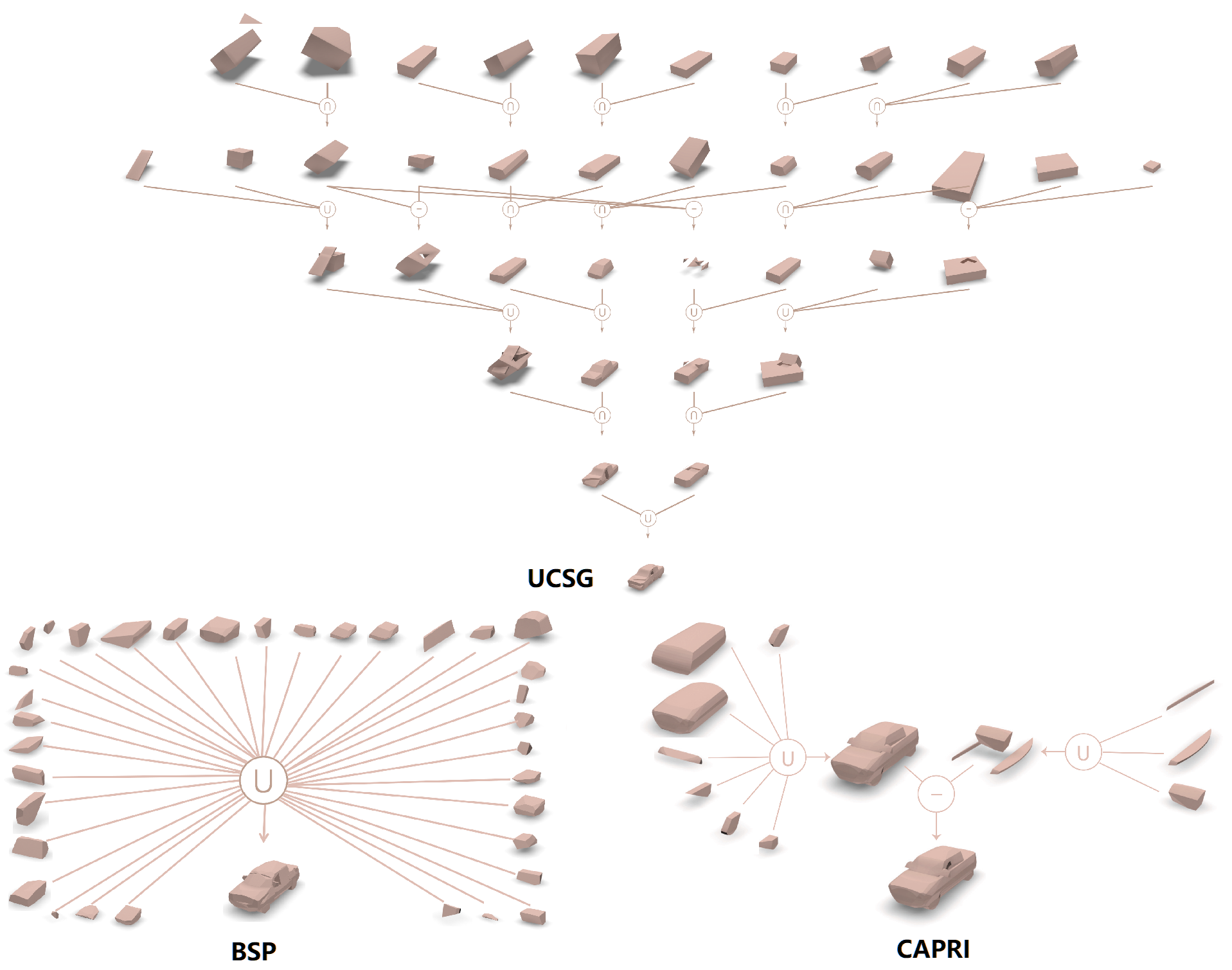}
  \caption{An example of CSG tree visual comparison on ShapeNet.}
  \label{fig:shapenet_tree_2}
\end{figure*}

\begin{figure*}
  \includegraphics[width=1\linewidth]{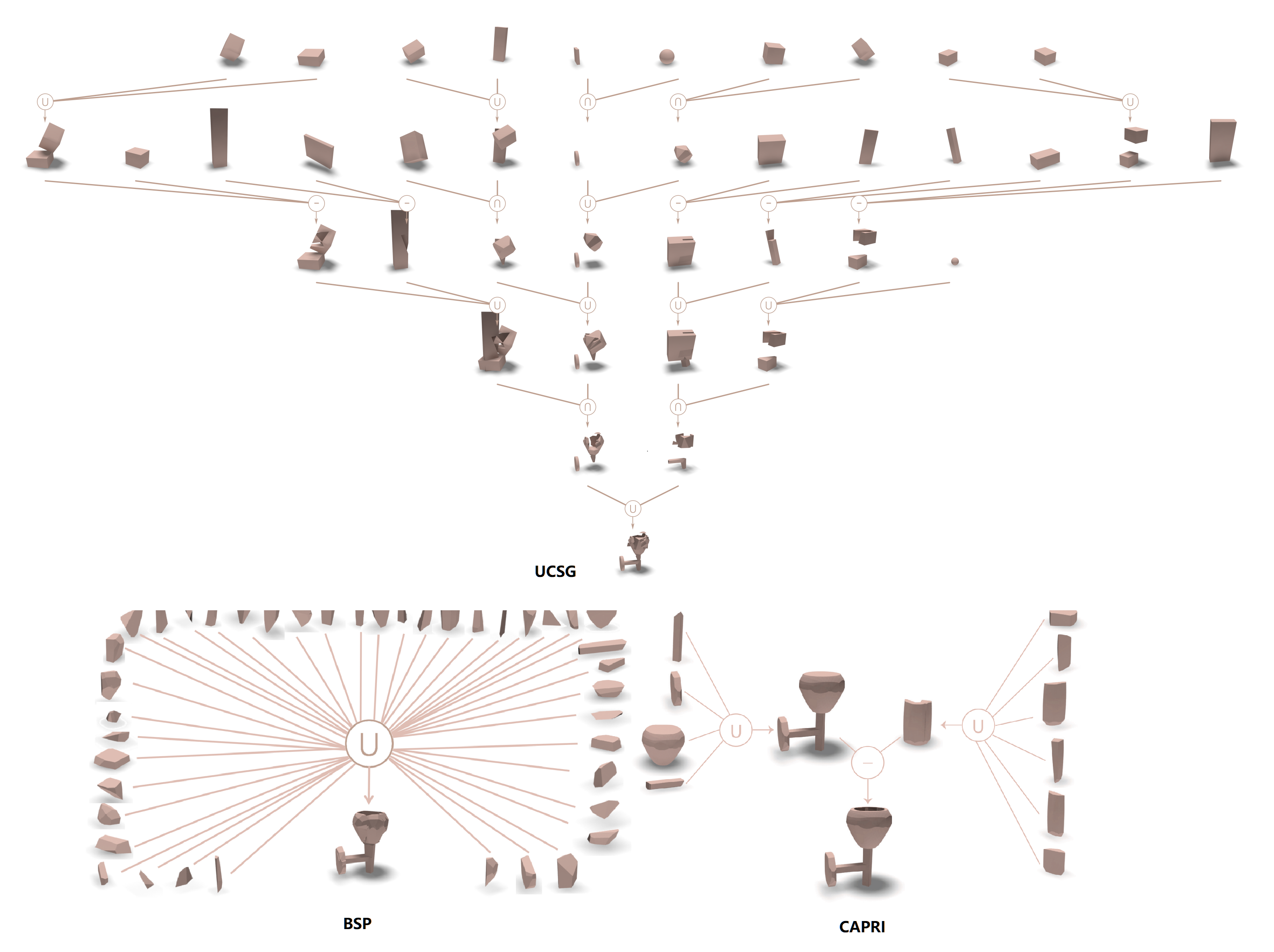}
  \caption{An example of CSG tree visual comparison on ShapeNet.}
  \label{fig:shapenet_tree_3}
\end{figure*}

\begin{figure*}
  \includegraphics[width=1\linewidth]{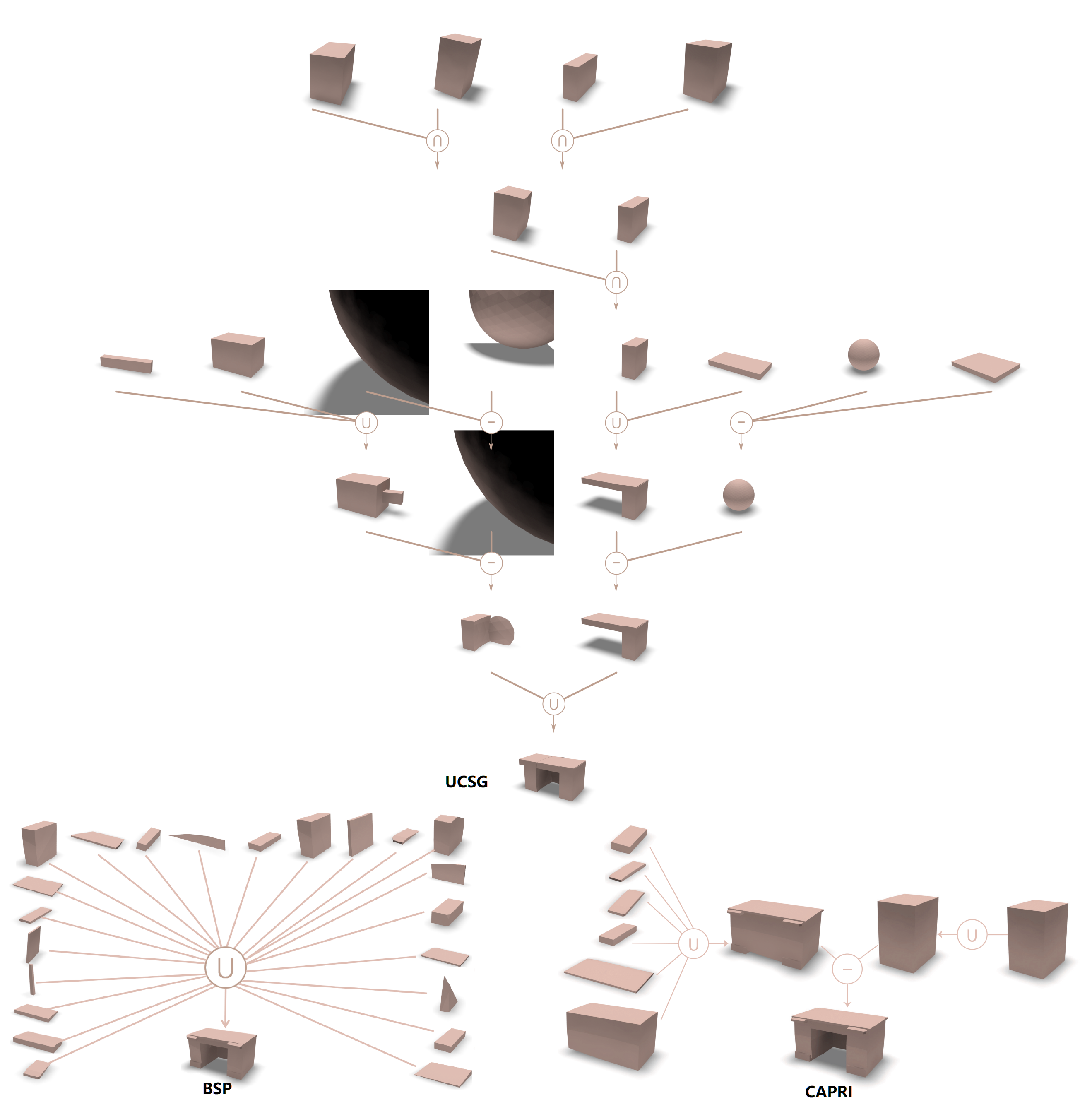}
  \caption{An example of CSG tree visual comparison on ShapeNet.}
  \label{fig:shapenet_tree_4}
\end{figure*}

We also provide CSG tree visual comparisons to BSP and UCSG, see Figures~\ref{fig:abc_tree_1}--\ref{fig:shapenet_tree_4}. 
Leaf nodes represent convex shapes in these trees. For BSP and CAPRI-Net, we do not show surface primitives for simplicity. UCSG does not use surface primitives but boxes and spheres, which are considered as convex shapes. According to the figure, the shapes produced by our method need fewer convex shapes in comparison with the other two methods, which makes our CSG tree more visually compact.

\vspace{-10pt}

\vspace{-10pt}

\vspace{-10pt}

\vspace{-10pt}

\vspace{-10pt}

\vspace{-10pt}

\vspace{-10pt}

\vspace{-10pt}

%% file: Tables/voxel.tex
\setlength{\tabcolsep}{4pt}
\begin{table}[h!]

\centering
 \begin{tabular}{| l | r | r | r |r |r |r |}
     \hline
 Methods & CD $\downarrow$ & NC $\uparrow$ & ECD $\downarrow$ & LFD $\downarrow$ \\
 \hline \hline
BSP w.o FT  & 0.75 & 0.83 & 12.01 & 2400.3 \\
UCSG w.o FT & 12.69 & 0.44 & 25.73 & 4538.3\\
Ours w.o FT & 0.56 & 0.84 & 12.78 & 2149.0\\
BSP         & 0.27 & 0.90 & 5.66 & 1149.7 \\
UCSG        & 0.18 & 0.89 & 5.89 & 1172.8 \\
Ours        & \textbf{0.15} & \textbf{0.91} & \textbf{4.27} & \textbf{750.3} \\
\hline
\end{tabular}
\caption{Comparing 3D reconstruction from voxels on ABC. }
    \label{tab:voxel_abc}
\end{table}

\setlength{\tabcolsep}{4pt}
\begin{table}[h!]

\centering
 \begin{tabular}{|l| r| r| r| r| r| r|}
     \hline
 Methods & CD $\downarrow$ & NC $\uparrow$ & ECD $\downarrow$ & LFD $\downarrow$\\
    \hline \hline
BSP w.o FT  & 0.56 & 0.84 & 4.41 & 2645.3 \\
UCSG w.o FT & 1.28 & 0.77 & 4.47 & 4405.5\\
Ours w.o FT & 0.33 & 0.85 & 10.55 & 2709.1\\
BSP         & 0.24 & \textbf{0.87} & 2.07 & 2189.9 \\
UCSG        & 0.68 & 0.84 & 4.15 & 3489.2 \\
Ours        & \textbf{0.20} & 0.85 & \textbf{2.06} & \textbf{1906.3} \\
\hline
\end{tabular}
\caption{Comparing 3D reconstruction from voxels on ShapeNet. }
    \label{tab:voxel_shapenet}
\end{table}

%% file: Tables/point.tex
\setlength{\tabcolsep}{3pt}

\begin{table}[t!]
\centering

\begin{tabular}{|l| r| r| r| r| r| r|}
    \hline
 Methods & CD $\downarrow$ & NC $\uparrow$& ECD $\downarrow$& LFD $\downarrow$ & \#P $\downarrow$ & \#C $\downarrow$\\
    \hline \hline
BSP-Net      & 0.19 & 0.94 & 2.91 & 621.7 & 180.80 & 14.94 \\
UCSG         & 1.43 & 0.88 & 11.56 & 2428.5 & - & 12.92\\
IM-Net$_{128}$ & \textbf{0.06} & 0.95 & 3.36 & \textbf{343.7} & - & - \\
SIREN$_{128}$  & 0.07 & \textbf{0.97} & \textbf{2.42} & 422.8 & - & -\\
Ours         &  0.09 &  0.94 & 2.54 & 500.5 & 70.26 & 5.30 \\
  \hline
\end{tabular}
\caption{Comparing 3D point cloud reconstruction on ABC dataset. Both IM-Net and SIREN are at $128^3$ resolution.}

\label{tab:pc_abc}

\end{table}

\setlength{\tabcolsep}{3pt}

\begin{table}[t!]
\centering

\begin{tabular}{|l| r| r| r| r| r| r|}
    \hline
 Methods & CD $\downarrow$ & NC $\uparrow$& ECD $\downarrow$& LFD $\downarrow$ & \#P $\downarrow$ & \#C $\downarrow$\\
    \hline \hline
BSP-Net        & 0.25 & 0.89 & 1.91  & 2475.7 & 205.80 & 18.04 \\
UCSG           & 3.40 & 0.80 & 5.45  & 5255.8 & -      & 12.78\\
IM-Net$_{128}$ & 2.06 & 0.86 & 3.63  & 2537.8 & -      & - \\
SIREN$_{128}$  & \bf 0.15 & \bf 0.92 & \bf 1.39  & 2232.0 & -      & -\\
Ours           & 0.21 & 0.87 & 1.70  & \bf 1876.1 & \bf 90.94  & \bf 8.90 \\
  \hline
\end{tabular}
\caption{Comparing 3D point cloud reconstruction on ShapeNet. Both IM-Net and SIREN are at $128^3$ resolution.}

\label{tab:pc_shapenet}

\end{table}